\documentclass{article}

\PassOptionsToPackage{numbers, compress}{natbib}

\usepackage[preprint]{neurips_2025}

\usepackage[utf8]{inputenc} 
\usepackage[T1]{fontenc}    
\usepackage{hyperref}       
\usepackage{url}            
\usepackage{bm}
\usepackage{booktabs}       
\usepackage{amsfonts}       
\usepackage{nicefrac}       
\usepackage{microtype}      
\usepackage{xcolor}         
\usepackage{enumitem}
\usepackage{adjustbox}
\usepackage{amsmath}
\usepackage{amssymb}
\usepackage{amsthm}
\usepackage{cleveref}
\usepackage{wrapfig}
\usepackage{stackengine}
\usepackage{comment}
\usepackage{mathtools}
\usepackage{xcolor}
\usepackage{longtable}
\usepackage{multirow}
\usepackage{graphicx}
\usepackage{subcaption}
\usepackage{colortbl}

\hypersetup{
    colorlinks,
    linkcolor={red!50!black},
    citecolor={blue!50!black},
    urlcolor={blue!80!black}
}
\theoremstyle{plain}
\newtheorem{definition}{Definition}
\newtheorem{theorem}{Theorem}
\newtheorem{lemma}{Lemma}

\title{Why Do More Experts Fail? A Theoretical Analysis of Model Merging}

\author{
  Zijing Wang \textsuperscript{\normalfont 1}\thanks{Equal contribution} \quad
  Xingle Xu \textsuperscript{\normalfont 1}\footnotemark[1] \quad
  Yongkang Liu \textsuperscript{\normalfont 1}\footnotemark[1] \quad
  Yiqun Zhang \textsuperscript{\normalfont 1}\footnotemark[1] \quad
  Peiqin Lin \textsuperscript{\normalfont 2} \quad
  \\
  {\bf Shi Feng} \textsuperscript{\normalfont 1} \quad
  {\bf Xiaocui Yang} \textsuperscript{\normalfont 1} \quad
  {\bf Daling Wang} \textsuperscript{\normalfont 1}\thanks{Corresponding author} \quad
  {\bf Hinrich Schütze} \textsuperscript{\normalfont 2} \quad
  \\
  \textsuperscript{1}Northeastern University, China \\
  \textsuperscript{2}CIS, LMU Munich; MCML, Germany \\
}

\begin{document}

\maketitle

\begin{abstract}
Model merging dramatically reduces storage and computational resources by combining multiple expert models into a single multi-task model. Although recent model merging methods have shown promising results, they struggle to maintain performance gains as the number of merged models increases. In this paper, we investigate the key obstacles that limit the scalability of model merging when integrating a large number of expert models.
First, we prove that there is an upper bound on model merging. Further theoretical analysis reveals that the limited effective parameter space imposes a strict constraint on the number of models that can be successfully merged. Gaussian Width shows that the marginal benefit of merging additional models diminishes according to a strictly concave function. This implies that the effective parameter space becomes rapidly saturated as the number of merged models increases. Furthermore, using Approximate Kinematics Theory, we prove the existence of a unique optimal threshold beyond which adding more models does not yield significant performance improvements. At the same time, we introduce a straightforward Reparameterized Heavy-Tailed method (RHT) to extend the coverage of the merged model, thereby enhancing its performance. Empirical results on 12 benchmarks, including both knowledge-intensive and general-purpose tasks, validate our theoretical analysis. We believe that these results spark further research beyond the current scope of model merging. The source code is in the Github repository: \url{https://github.com/wzj1718/ModelMergingAnalysis}.
\end{abstract}
\section{Introduction}
General Artificial Intelligence is the ultimate goal pursued by researchers. Model merging offers a promising solution by integrating multiple task-specific expert models into a unified multi-task model. By combining the capabilities of diverse expert models, a merged system can handle a broader range of tasks and adapt more effectively to complex problems. The most direct approach involves performing arithmetic merging~\cite{yadav2023ties,yu2024language}, which combines multiple model parameters through mathematical operations to enhance the model's multi-task capabilities, such as weighted averaging. Since the parameter subspaces of different experts conflict, these arithmetic merging methods may lead to the collapse of the merged parameter space.
In order to avoid conflicts in parameter spaces among different experts, the orthogonal methods reduce interference of inconsistent parameters by merging the decomposed vertical parameters~\cite{po2024orthogonal,choi2024revisiting}.
Merging only mutually orthogonal parameters, which may result in the loss of crucial parameters.
Recently, researchers have proposed using evolutionary algorithms for model merging, significantly enhancing the merging performance~\cite{akiba2025evolutionary,zhang2025nature}.

These merging methods have achieved landmark performance, but they have limitations in model merging—specifically, only a small number of experts can be combined. Our preliminary experiments find that the performance of the current most advanced model merging method reaches saturation after fusing at most six models (e.g., the maximum number of Model Swarms~\cite{feng2024model} merging is about four, and the maximum number of GENOME~\cite{zhang2025nature} merging is approximately six).  Although some classical results~\cite{yadav2024matters,tao2024unlocking} suggest the presence of a saturation effect in model merging, the reasons behind it are unexplored. 

To this end, we leverage high-dimensional geometry~\cite{vershynin2015estimation} and the Approximate Kinematics Theory~\cite{amelunxen2014living} to investigate the underlying causes of the saturation phenomenon in model merging. First, we theoretically analyze the evolution of the parameter space of the merged model as the number of experts increases. 
We find that as the number of experts increases, the Gaussian Width of the parameters no longer grows, indicating that the effective parameter space of the merged model gradually saturates, leading to a performance bottleneck.
Furthermore, leveraging Approximate Kinematics Theory~\cite{amelunxen2014living}, we derive an optimal upper bound for model merging. We also observe that the effective parameter space of the merged model is highly sparse, resulting in limited coverage. To address this, we propose a simple Reparameterized Heavy-Tailed (RHT) method, which enhances the model’s parameter space coverage by amplifying the heavy-tailed distribution, thereby improving performance.
Experiments on both knowledge-intensive and general-purpose tasks provide extensive validation of our theory. Our main contributions and findings are summarized as follows:
\begin{itemize}[leftmargin=*]
  \item We prove that as the number of expert models increases, the effective parameter space of the model rapidly saturates, leading to diminishing returns in performance. 
  \item We prove the existence of an upper bound for model merging and provide its analytical expression, highlighting performance limitations caused by parameter redundancy and offering theoretical guidance for optimizing expert model merging.
  \item We propose a simple Reparameterized Heavy-Tailed (RHT) method to enhance the coverage of the merged model by extending its Heavy-Tailed distribution.
  \item Experiments on both knowledge-intensive and general-purpose tasks validate the correctness and effectiveness of our theories and methods.
\end{itemize}

\section{Theory}
Model merging refers to the integration of multiple task-specific expert models into a unified multi-task model. 
To formally describe the merging process, let $\theta_0 \in \mathbb R^{d}$ denote the weights of the pre-trained model. Getting experts with LoRA is a popular method. Thus, we assume that the experts are obtained through LoRA fine-tuning. Let \{$\theta_1, \theta_2, \cdots, \theta_M$\} represent the LoRA expert parameters that need to be merged, where $M$ represents the number of experts. 


We prove that there exists an upper bound to model merging and provide a theoretical adaptive termination condition (\Cref{Upper_Bound_o_Merge_Merging}). To further investigate the cause of this upper bound, we analyze the diminishing marginal returns of model merging using Gaussian Width (\Cref{marginal-effects-of-gaussian-width}) and examine the saturation of the merged model’s effective parameter space through the Approximate Kinematics Theory (\Cref{parameter-redundancy-effects-approximate-kinematics}). Based on these insights, we propose a simple Reparameterized Heavy-Tailed method to improve the coverage of the merged model (\Cref{reparameterized-heavy-tailed-method}).
\begin{theorem}[Upper Bound of Model Merging] 
\label{Upper_Bound_o_Merge_Merging}
As the number of merging experts increases, the variance of the combined model approaches a constant and the performance of the model approaches saturation(proof in the \Cref{proof:Upper_Bound_o_Merge_Merging}).

A large number of experiments have shown that the incremental parameter distribution of LoRA experts conforms to the normal distribution (\Cref{sub:lora}), so we assume that $\theta_i \sim \mathcal{N}(0, \sigma_i^2 I)$, according to the linear combination property of Gaussian random variables, the parameter distribution after fusion is $ \theta^k = \sum_{i=1}^n \alpha_i \,\theta_i$, where the weight coefficient $\alpha_i$ satisfies the constraint $\sum_{i=1}^n \alpha_i = 1, \alpha_i \ge 0$. Considering the correlation between experts, the combined variance after combining different experts is expressed as
\begin{equation}
   \sigma_{\mathrm{merge}}^2
= \sum_{i=1}^n \alpha_i^2\,\sigma_i^2
  + \sum_{i \neq j} \alpha_i \alpha_j\,\rho_{ij}\,\sigma_i\,\sigma_j,
\end{equation}
where $\rho_{ij}$ is the correlation coefficient. When there is a $\rho_{ij}$ between the expert models, the combined variance has a lower bound. We assume that the variances of all experts are equal $\sigma_i^2 = \sigma^2$, and the correlation coefficients between experts are equal $\rho_{ij} = \rho$:
\begin{equation}
    \sigma_{\mathrm{merge}}^2 = \sigma^2 \Bigl(\rho + (1 - \rho)\sum_{i=1}^n \alpha_i^2\Bigr).
\end{equation}
When the number of experts merged $n \to \infty$, the variance after merging tends to:
\begin{equation}
    \lim_{n \to \infty} \sigma_{\mathrm{merge}}^2 = \sigma^2 \rho.
\end{equation}
This indicates that there is a theoretical lower bound $\sigma^2 \rho$ for the merge variance. To ensure that each expert reduces the variance by at least $ \Delta $, the upper bound of the number of merged experts is:
\begin{equation}
    n\le\frac{\sigma^2(1-\rho)}{\Delta}.
\end{equation}
Our theoretical framework demonstrates that there is an upper bound to the number of experts $n$ that can be effectively merged and that indefinitely increasing the number of merged experts does not consistently lead to performance improvements. When high-performance experts are merged (i.e., $\Delta$ is large), a smaller number of experts $n$ is sufficient to achieve strong performance. Enhancing the orthogonality constraint between experts by regularizing the value of $\rho$ can potentially improve the performance of the merging model. To determine whether it is necessary to continue merging models, we introduce an adaptive termination condition $\Delta = \mathbb{E}\left[\rho_{i-1}^2-\rho_i^2\right]$. If the variance reduction achieved by incorporating a new model is negligible or falls below a specified threshold, the merging process can be terminated.
\end{theorem}
\subsection{Marginal Effects of Gaussian Width in Parameter Subspace}
\label{marginal-effects-of-gaussian-width}
\begin{theorem}[Diminishing Marginal Effects in Model Merging]\label{theorem:1} As the number of expert models $M$ increases, the addition of new experts continues to expand the dimensionality of the parameter space. However, the marginal effects of each new dimension on the Gaussian Width diminish progressively, leading to the saturation of the performance of expert model merging. For the number of experts $M$, the Gaussian Width becomes (Proof in the \Cref{sub:gaussian width}):
\begin{equation}
w(S_M) \approx \sqrt{2\epsilon \cdot \sum_{i=1}^{M} \frac{1}{\lambda_i}},
\end{equation}
where $\lambda_i$ is the $i$-th eigenvalue of $H$.
The marginal contribution of adding the $ M $-th expert is:
\begin{equation}
\Delta w_M = w(S_M) - w(S_{M-1}) = \sqrt{2\epsilon \cdot \sum_{i=1}^{M} \frac{1}{\lambda_i}} - \sqrt{2\epsilon \cdot \sum_{i=1}^{M-1} \frac{1}{\lambda_i}}.
\end{equation}
Since the square root function is concave, the marginal gain decreases as $ M $ increases:
\begin{equation}
\Delta w_M > \Delta w_{M+1}.
\end{equation}\end{theorem}
Thus, diminishing marginal return arises from the concavity of the square root function, leading to progressively smaller contributions from each additional expert to the overall Gaussian Width.
\subsection{Parameter Redundancy Effects via Approximate Kinematics}
\label{parameter-redundancy-effects-approximate-kinematics}
\begin{theorem}[Parameter Redundancy and Expert Model Merging Performance]\label{theorem:2} As the number of merged expert models $M$ increases,  the number of non-zero parameters $k$ in the network gradually grows. When parameter redundancy exceeds a certain threshold, it becomes impossible to maintain the loss within the sublevel set, resulting in a decline in model performance. Specifically, when the number of non-zero parameters $k$ satisfies the following inequality: \begin{equation}
    k \leq D -  \sum_{i=1}^{D-k} \frac{ r_i^2}{\|\theta^* - \theta^k\|_2^2 + r_i^2}.
\end{equation}
Once this threshold is exceeded, performance degradation becomes inevitable(Proof in the \Cref{sub:Parameter Redundancy}).
\end{theorem}

\subsection{Reparameterized Heavy-Tailed Method}\label{reparameterized-heavy-tailed-method}

Based on the experimental observations in \Cref{sub:Experiments}, we find that the parameters of the merged multi-expert model, $\mathbf{w} \in \mathbb{R}^d$, approximately follow a multivariate Gaussian distribution $
\mathcal{N}(\boldsymbol{\mu}, \boldsymbol{\Sigma})$, where $\boldsymbol{\mu} \in \mathbb{R}^d$ is the mean vector and $\boldsymbol{\Sigma} \in \mathbb{R}^{d \times d}$ is the covariance matrix (weight distribution histograms are provided in~\ref{fig:merge weight}). For simplicity, we assume $\boldsymbol{\Sigma} = \sigma^2 \mathbf{I}$, and define a two-step transformation:
\begin{enumerate}[leftmargin=*]
    \item Gaussian difference: $\mathbf{w}' = \mathbf{w} - \mathbf{g}, \quad \mathbf{g} \sim \mathcal{N}(\boldsymbol{\mu}, \sigma_g^2 \mathbf{I})$. \item Component-wise nonlinear amplification: $\mathbf{w}'' = T(\mathbf{w}'), \quad T: \mathbb{R}^d \to \mathbb{R}^d$.
\end{enumerate}
\begin{theorem}[Difference of Two Independent Gaussian Random Vectors]\label{theorem:4}
The difference of two independent Gaussian random vectors remains Gaussian. Let $\mathbf{w} \sim \mathcal{N}(\boldsymbol{\mu}, \sigma^2 \mathbf{I})$ and $\mathbf{g} \sim \mathcal{N}(\boldsymbol{\mu}, \sigma_g^2 \mathbf{I})$ be independent random vectors. Then, their difference $\mathbf{w}' = \mathbf{w} - \mathbf{g}$  
 follows a Gaussian distribution $\mathbf{w}' \sim \mathcal{N}(\mathbf{0}, (\sigma^2 + \sigma_g^2) \mathbf{I})$ (Proof in the \Cref{sub:Difference of Gaussian Distributions}).
\end{theorem}
\begin{theorem}[Nonlinear Transformation Induces Heavy-Tailed Distributions]\label{theorem:5} Let $\mathbf{w}' \sim \mathcal{N}(\mathbf{0}, (\sigma^2 + \sigma_g^2) \mathbf{I})$ be a zero-mean multivariate Gaussian distribution. Consider a nonlinear transformation $T: \mathbb{R}^d \rightarrow \mathbb{R}^d$, where for each component $i$, the transformation is defined as  
\begin{equation}
T(w'_i) = \mathrm{sign}(w'_i) \cdot |w'_i|^{\gamma} \cdot \left(1 + \alpha \cdot e^{-\beta |w'_i|}\right),
\end{equation}
with parameters $0 < \gamma < 1$, $\alpha > 0$, and $\beta > 0$. Then the transformed random vector $\mathbf{w}'' = T(\mathbf{w}')$
follows a heavy-tailed distribution, whose marginal probability density function for each component $\mathbf{w}''_i$ satisfies  
\begin{equation}
p_{\mathbf{w}''_i}(y_i) \propto |y_i|^{\frac{1}{\gamma} - 1} \exp\left(-\frac{|y_i|^{\frac{2}{\gamma}}}{2(\sigma^2 + \sigma_g^2)}\right).
\end{equation}
Furthermore, for sufficiently large $|y_i|$, the tail behavior of the cumulative distribution function satisfies  
\begin{equation}
P(|W''_i| > |y_i|) \sim |y_i|^{-\kappa},
\end{equation}
where the tail exponent is given by $\kappa = \frac{1}{\gamma}$. As $\gamma \to 0$, the distribution exhibits heavier tails, and compared to the original Gaussian distribution, $\mathbf{w}''$ has a higher probability of extreme values. (Proof in the ~\Cref{sub:Nonlinear Transformation}).
\end{theorem}
\begin{theorem}[Heavy-Tailed Distributions Enhance Model Coverage]\label{theorem:6} Let the function space defined by a neural network be $\mathcal{F}_{\mathbf{w}} = \{ f_{\mathbf{w}}(\mathbf{x}) : \mathbf{w} \in \mathcal{W} \}$, where $\mathcal{W}$ is the parameter space, and $f_{\mathbf{w}}$ is the neural network parameterized by $\mathbf{w}$. Define the coverage of the function space as
\begin{equation}
\mathcal{C}(\mathcal{F}) = \int_{\mathcal{X}} \big| \{ f(\mathbf{x}) : f \in \mathcal{F} \} \big| \, d\mathbf{x},
\end{equation}
where $\mathcal{X}$ is the input space. If the original distribution of parameters $\mathbf{w}$, denoted $p_{\mathbf{w}}$, is Gaussian, and after a transformation (subtracting a Gaussian and applying a nonlinear amplification to the residual parameters) the distribution $p_{\mathbf{w}''}$ becomes heavy-tailed, then under the parameter-to-function mapping $\Phi : \mathcal{W} \to \mathcal{F}$, the coverage of the transformed model $\mathcal{C}_2$ is strictly greater than that of the original model $\mathcal{C}_1$, i.e.,
\begin{equation}
\mathcal{C}_2 = \int_{\mathcal{W}} \big|\det(J_{\Phi}(\mathbf{w}))\big| \, p_{\mathbf{w}''}(\mathbf{w}) \, d\mathbf{w} > \int_{\mathcal{W}} \big|\det(J_{\Phi}(\mathbf{w}))\big| \, p_{\mathbf{w}}(\mathbf{w}) \, d\mathbf{w} = \mathcal{C}_1,
\end{equation}
where $J_{\Phi}(\mathbf{w})$ is the Jacobian matrix of the mapping $\Phi$ at $\mathbf{w}$, and $|\det(J_{\Phi}(\mathbf{w}))|$ denotes the local volume change ratio from the parameter space to the function space (Proof in the ~\Cref{sub:Heavy-Tailed Distributions Expand the Model Function Space}).
\end{theorem}

\section{Experiments}\label{sub:Experiments}

All fine-tuning with backpropagation experiments follow convention and use Adam as optimizer. The detailed settings, including hyperparameters, are reported in \Cref{app:implementation_details}.
All experiments below use datasets detailed in \Cref{app:datasets}. We also discuss the limitations of this paper (\Cref{sec:limitations}). 
\subsection{Upper Bound for Model Merging}
\label{sub:upper_bound_for_model_merging}

As shown in Table \ref{tab:Performance of GENOME and Model}, experiments conducted on the $D_{\text{gend}}$ task with both GENOME and Model Swarms reveal that, although both methods from the original papers involve merging $10$ expert models, our experiments with $2$, $4$, $6$, $8$, and $10$ LoRA models indicate that the optimal performance of model merging is not achieved with $10$ LoRA models, as performance reaches saturation earlier.

As discussed in~\Cref{theorem:1}, as the number of expert models increases, the Gaussian Width of the parameter subspace exhibits diminishing returns and eventually reaches saturation. This phenomenon arises from the fact that the newly added experts occupy directions in the Hessian curvature space that progressively shift toward low-curvature regions, leading to a diminishing marginal contribution to the expansion of the model's representational capacity. Our empirical results align closely with this theoretical prediction. Specifically, although each additional expert expands the parameter subspace, earlier experts have already covered the primary high-curvature directions in the Hessian space. As a result, subsequent experts predominantly contribute to low-curvature directions, which correspond to smaller eigenvalues, and thus have limited capacity to adjust the loss function, leading to a gradual reduction in overall performance improvement.

\begin{wrapfigure}[]{r}{0.4\textwidth}
    \centering
    \includegraphics[width=0.4\textwidth]{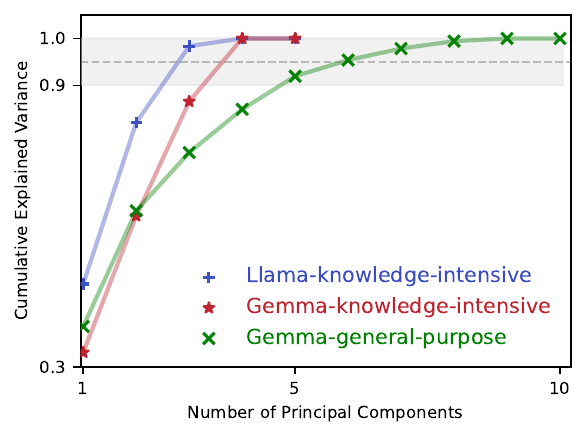}
    \caption{Cumulative variance comparison across different expert models.}
    \label{fig:1}
    \vspace{-0.5em}
\end{wrapfigure}

To further verify this phenomenon, we perform principal component analysis (PCA)~\cite{wold1987principal} on the weights of various expert models used in the experiments (see Figure \ref{fig:1}). The results indicate that the number of principal components explaining approximately 95\% of the total variance closely corresponds to the number of expert models at which the model performance peaks. This suggests that while the number of activated parameter dimensions (i.e., explained variance) continues to increase as more experts are added, the actual performance of the model no longer improves and may even degrade.

According to the analysis of~\Cref{theorem:2}, this performance degradation can be attributed to the increase in parameter redundancy. As the number of experts increases, the number of non-zero parameters $k$ also increases, causing $k$ to exceed the theoretical upper limit, resulting in performance degradation.

In summary, our empirical findings strongly align with the theoretical analysis: expert model merging can effectively enhance performance within a certain range, but as the number of experts increases, the marginal benefit gradually decreases, and performance is ultimately limited by parameter redundancy.

\begin{table*}[htbp]
        \caption{Performance of GENOME and Model Swarms with 2-10 LoRA Fusion on $D_{\text{gend}}$ Corpus. The results are averaged over 5 runs with different random seeds.}
	\begin{adjustbox}{max width=\textwidth, center}
		\includegraphics[width=\textwidth]{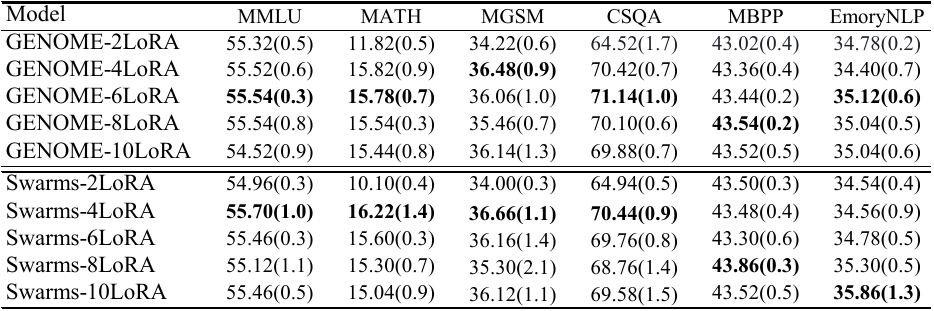}
	\end{adjustbox}
	\label{tab:Performance of GENOME and Model}
    \vspace{-1.5em}
\end{table*}



\begin{figure}[htbp]
    \centering
    \begin{minipage}[t]{0.52\textwidth}
        \centering
        \includegraphics[width=\linewidth]{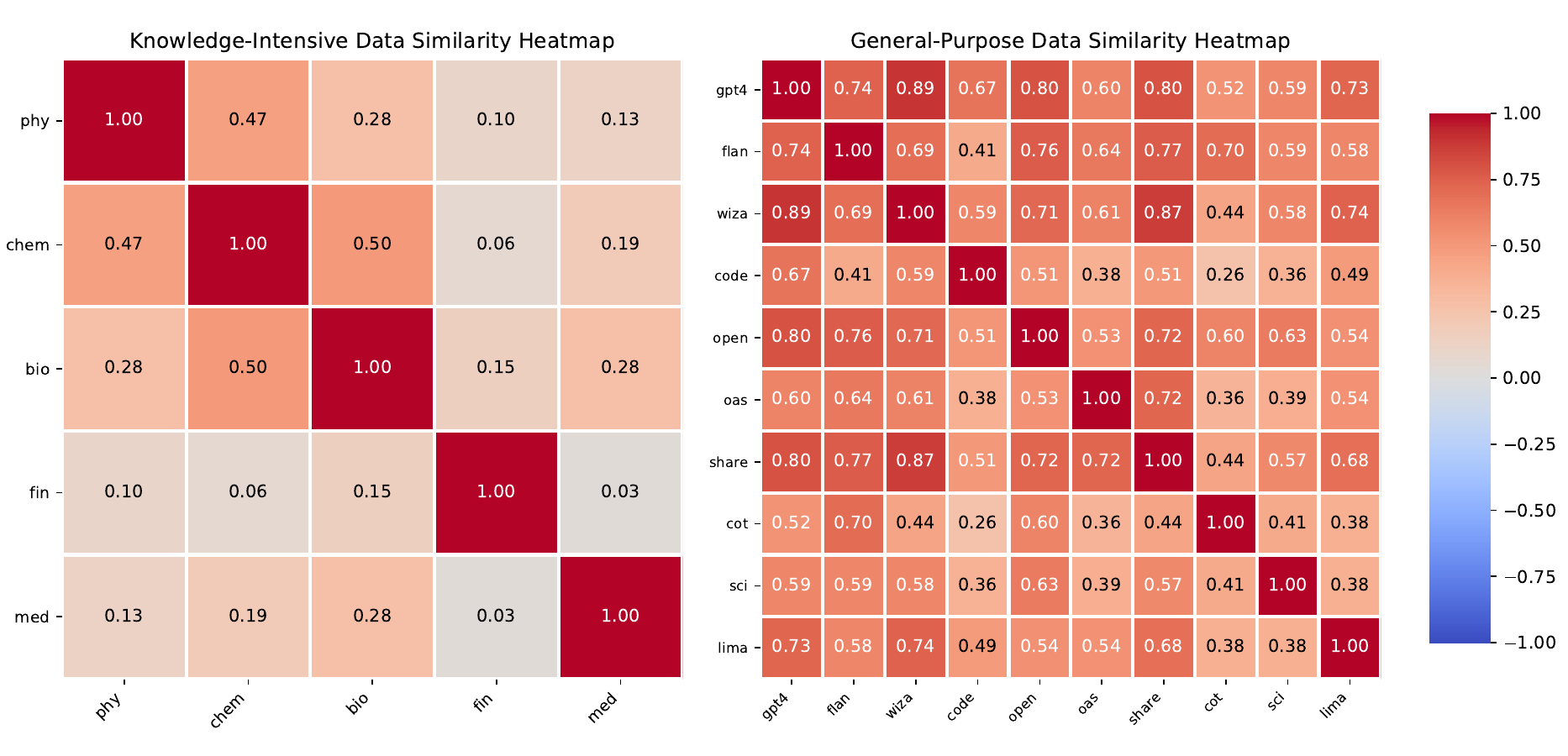}
        \caption{Heatmap of Cosine Similarities Between Sentence Embeddings of $D_{\text{knowd}}$ and $D_{\text{gend}}$.}
        \label{fig:heatmap}
    \end{minipage}
    \hfill
    \begin{minipage}[t]{0.455\textwidth}
        \centering
        \includegraphics[width=\linewidth]{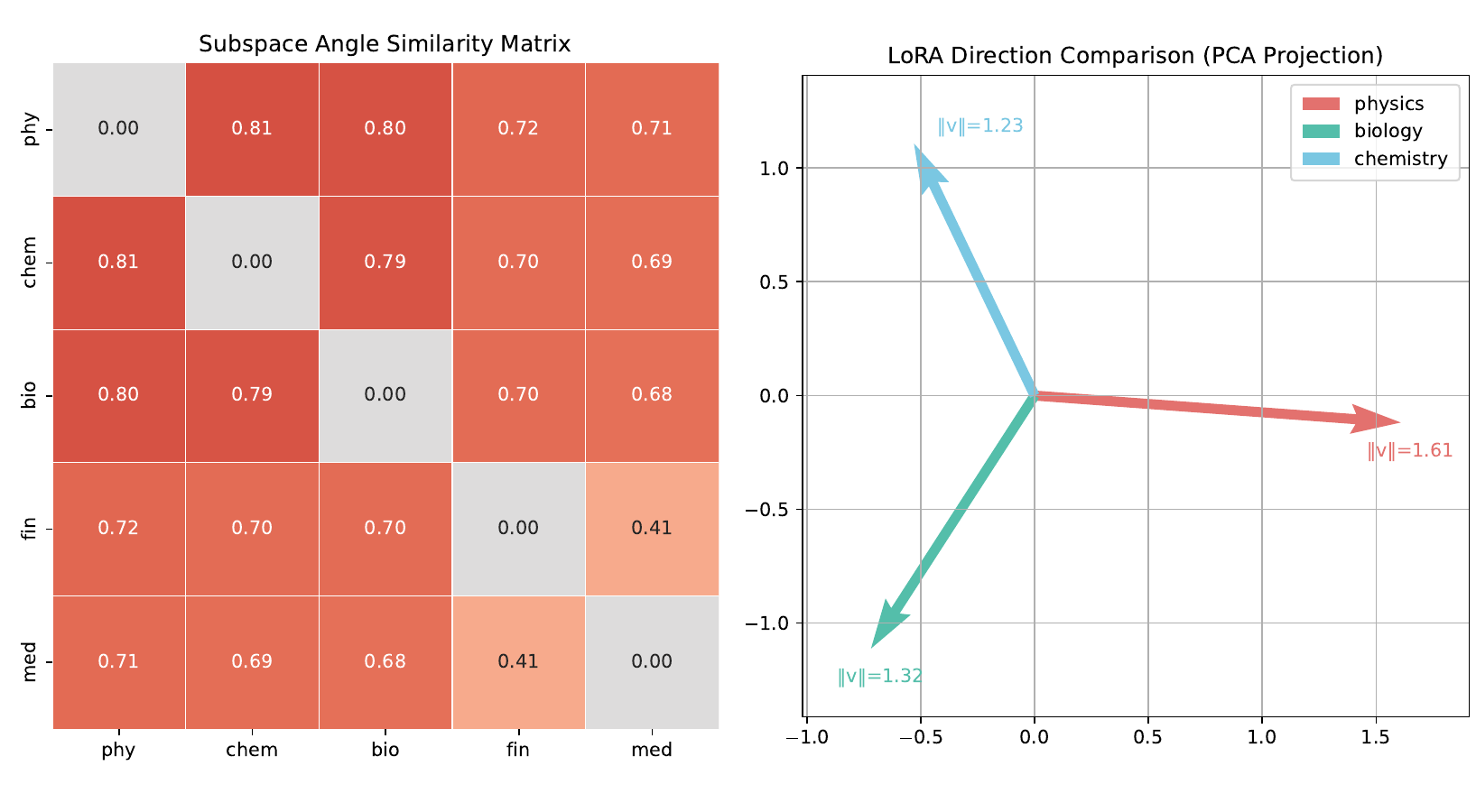}
        \caption{Analysis of domain model subspace orthogonality and PCA projections.}
        \label{fig:three_panel}
    \end{minipage}
    \vspace{-0.5em}
\end{figure}

\subsection{Impact of Domain Similarity on Model Merging}
\label{sub:impact_of_domain_similarity_on_model_merging}

Figure \ref{fig:heatmap} illustrates the cosine similarity between the embeddings of the training corpora $D_{\text{knowd}}$ and $D_{\text{gend}}$, clearly reflecting the differences in correlation between these two types of data. We conduct model merging experiments based on these two corpora with differing correlation levels, and the results are shown in Tables \ref{tab:Performance of GENOME and Model} and \ref{tab:one}. In both settings, the merging performance exhibits saturation.

According to~\Cref{Upper_Bound_o_Merge_Merging}, blindly increasing the number of expert models does not always lead to performance improvements. Enhancing the quality and diversity of individual expert models is often more effective than simply increasing the number of experts. \Cref{tab:pairwise LoRA fusion experiments} presents the results of merging two expert models from either different domains or the same domain. Experiments on two test sets in the physics domain indicate that merging expert models from different domains yields better performance than merging those from the same domain. \Cref{Upper_Bound_o_Merge_Merging} further states that the upper bound on the number of models that can be effectively merged is primarily constrained by the correlation between experts. The higher the correlation, the stricter the upper bound. Therefore, when merging expert models, prioritizing combinations of experts with lower correlation tends to achieve better performance gains.

\begin{table*}[!h]
        \caption{Zero-shot performance comparison of different LoRA fusion settings on Gemma-2-2B-it (top) and LLaMA3.1-8B-Instruct (bottom) models across various domain-specific tasks. ``Single'' to a LoRA model trained on $D_{\text{knowd}}$, ``3-LoRA'', ``4-LoRA'', and ``5-LoRA'' correspond to the first 3, 4, and 5 items in the sequence of ``physics, chemistry, biology, finance, and medicine''.}
	\begin{adjustbox}{max width=\textwidth, center}
		\includegraphics[width=\textwidth]{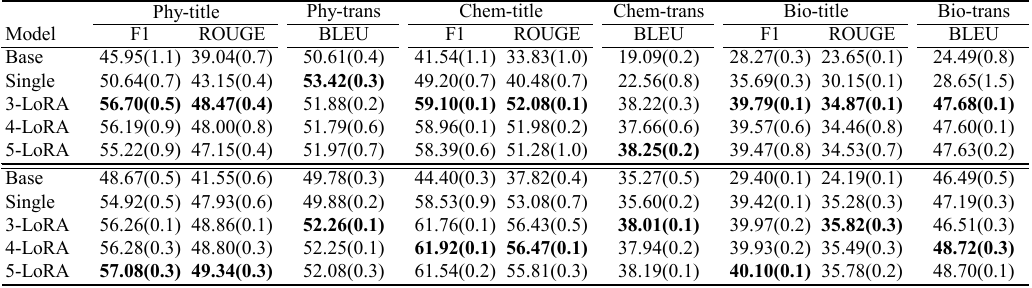}
	\end{adjustbox}
	\vspace{-1.5em}
	\label{tab:one}
\end{table*}

\begin{table*}[!h]
        \caption{Performance of pairwise LoRA fusion experiments across domains.}
	\begin{adjustbox}{max width=\textwidth, center}
		\includegraphics[width=\textwidth]{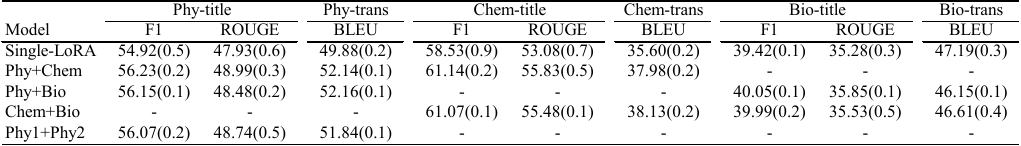}
	\end{adjustbox}
    \vspace{-1em}
	\label{tab:pairwise LoRA fusion experiments}
\end{table*}
Singular Value Decomposition (SVD)~\cite{klema1980singular} is a widely used matrix factorization technique for extracting principal components from data matrices. Based on SVD, we analyze the representation differences among five models in $D_{\text{knowd}}$, focusing on the similarity between the model subspaces from different domains. We measure the principal angles between these subspaces. As shown in the left part of Figure \ref{fig:three_panel}, the principal angles between the physics, chemistry, and biology model subspaces are close to 90 degrees, indicating near orthogonality and minimal parameter interference. In contrast, the principal angles between the finance and medical subspaces are smaller, revealing a significant overlap. This overlap reflects partial shared features but may also contain conflicting domain-specific information, causing optimization conflicts and performance degradation during fusion. The 4LoRA and 5LoRA merging experiments in Table \ref{tab:one} further confirm this phenomenon: adding medical LoRA to 4LoRA results in poor performance. This indicates that the coupling between the finance and medical model subspaces, driven by domain differences, induces negative interference that limits fusion effectiveness.

To more precisely quantify the differences between domain models, we perform PCA analysis to examine the representations of physics, chemistry, and biology domains in the reduced-dimensional space (see the right part of Figure \ref{fig:three_panel}). The PCA projections show that the vectors from these three domains point in different directions along the first two principal components, revealing significant differences in variation patterns. Due to the approximate orthogonality of their subspaces, the vectors exhibit minimal overlap, effectively reducing interference during fusion and ensuring stability and independence in parameter integration. The relative balance in vector magnitudes indicates that each domain contributes comparably to the fusion, which facilitates overall improvement in the fused model’s performance.

Table \ref{tab:pairwise LoRA fusion experiments} compares the performance of single LoRA models against pairwise fusion of approximately orthogonal LoRA models. The results demonstrate that fused pairs consistently outperform single-domain expert models. The approximate orthogonality of the subspaces ensures relative independence among the update directions of each adapter, effectively minimizing parameter interference during fusion, thereby promoting effective integration of knowledge across domains and enhancing overall model performance.

\begin{figure}[!h]
  \centering
  \includegraphics[width=0.7\linewidth]{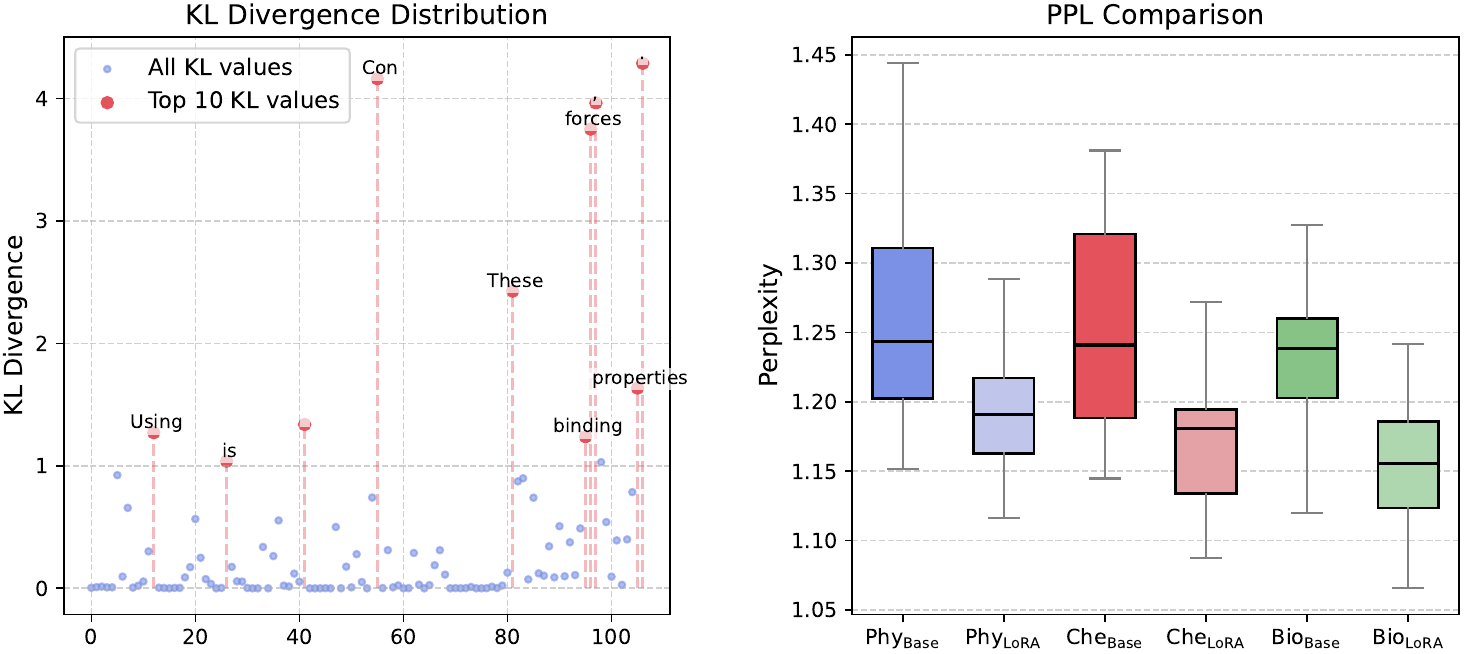}
  \caption{KL divergence distribution and perplexity comparison. (Left) The KL divergence distribution between the base and LoRA fine-tuned models, highlighting the Top 10 KL values (marked in red) corresponding to domain-independent tokens, indicating minimal adjustment to the overall model output. (Right) Perplexity comparison across domains (physics, chemistry, and biology) for both the base and LoRA fine-tuned models, showing a reduction in perplexity for the LoRA models, suggesting improved response accuracy and stability within each domain.}
  \label{fig:kl}
\end{figure}

\begin{figure}[!h]
  \centering
  \includegraphics[width=\linewidth]{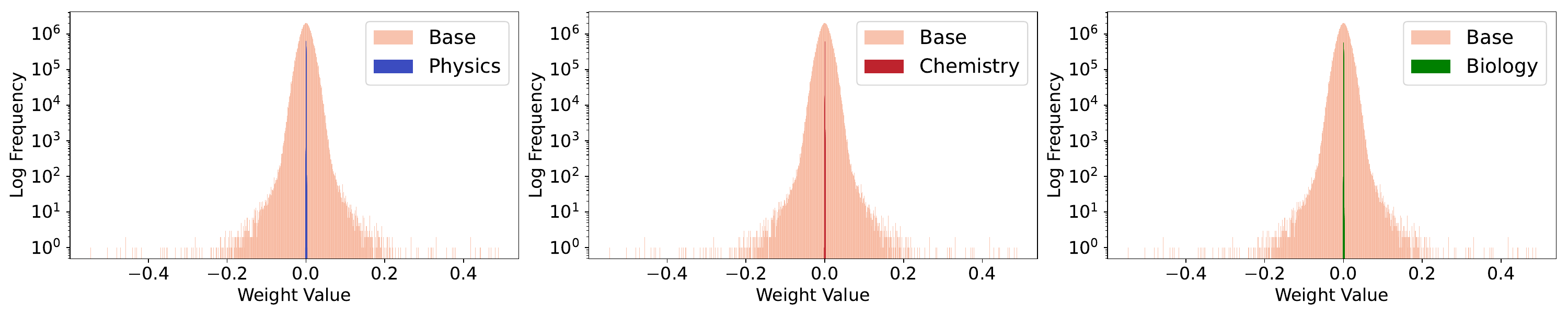}
  \caption{Weight distributions of the base model vs. LoRA fine-tuning on different domains. The histograms show the log frequency of weight values.}
  \label{fig:histogram}
\end{figure}
\subsection{The Limitations of LoRA Experts on Model Merging}\label{sub:lora}
Figure \ref{fig:histogram} presents the weight distribution histograms of the base model and the LoRA fine-tuned model across different domains. The results show that the LoRA weights exhibit a highly sparse distribution, indicating that the adjustments made to the original model parameters are extremely limited. Further quantification through SVD reveals a significantly long-tailed distribution of the LoRA weight singular values: only 0.195\% of the singular values fall within the range of $e^{-1}$ to $e^{-2}$, while the remaining singular values are below $e^{-9}$. This suggests that most of the parameter changes in LoRA fine-tuning are concentrated in a few directions, with minimal contribution to the overall behavior of the model.

To validate the effect of these parameter space constraints on model outputs, we perform a token-level KL divergence analysis. Given an input sequence $x_{\mathrm{input}}$ and an answer sequence $x_{\mathrm{ans}}$, we construct a complete context by concatenating them: $x_{1:t_{\mathrm{ans}}} = [x_{\mathrm{input}}; x_{\mathrm{ans},1:t_{\mathrm{ans}} - |x_{\mathrm{input}}|}]$, where $t_{\mathrm{ans}} \in [|x_{\mathrm{input}}| + 1, |x_{\mathrm{input}}| + |x_{\mathrm{ans}}|]$. We then compare the output distribution differences between the base model $\theta_{\mathrm{base}}$ and the LoRA model $\theta_{\mathrm{LoRA}}$:
\begin{equation}
\mathcal{L}_{\mathrm{KL}}(\theta_{\mathrm{base}}, \theta_{\mathrm{LoRA}})
= D_{\mathrm{KL}}\bigl(P_{\theta_{\mathrm{base}}}(\,\cdot\,\mid x_{1:t_\mathrm{ans}}) \parallel P_{\theta_{\mathrm{LoRA}}}(\,\cdot\,\mid x_{1:t_\mathrm{ans}})\bigr).
\end{equation}
The specific calculation is performed on a per-token basis:
\begin{equation}
D_{\mathrm{KL}}
= \sum_{v\in\mathcal{V}}
P_{\theta_{\mathrm{base}}}(v\mid x_{1:t_{\mathrm{ans}}})
\log\!\Bigl(\frac{P_{\theta_{\mathrm{base}}}(v\mid x_{1:t_{\mathrm{ans}}})}
{P_{\theta_{\mathrm{LoRA}}}(v\mid x_{1:t_{\mathrm{ans}}})}\Bigr).
\end{equation}
Here, $P_{\theta_{\mathrm{base}}}(v \mid x_{1:t_{\mathrm{ans}}})$ and $P_{\theta_{\mathrm{LoRA}}}(v \mid x_{1:t_{\mathrm{ans}}})$ are the logits assigned by the base model and the LoRA model, respectively, to token $v$ in the vocabulary $\mathcal{V}$, given the prefix $x_{1:t_{\mathrm{ans}}}$.

The result on the left of Figure \ref{fig:kl} shows that significant KL divergence differences (Top 10) mainly appear on domain-independent tokens. This suggests that LoRA fine-tuning does not reconstruct the output distribution by introducing new knowledge, but rather improves the original model’s performance on specific tasks by adjusting a small number of parameters.

To quantify the improvement in model performance, we conduct perplexity (PPL) comparison experiments across the domains of physics, chemistry, and biology. For each domain, we randomly select 40 questions and generate responses using both the LLaMA3.1-8B-Instruct and the LoRA fine-tuned model for that specific domain. As shown in the right part of Figure \ref{fig:kl}, the perplexity of the LoRA model is generally lower than that of the basic model, indicating that LoRA fine-tuning enhances the model's response accuracy in the target domain while also increasing output stability. The experimental results suggest that LoRA fine-tuning essentially refines the model’s inherent capabilities through optimization of a low-dimensional manifold in the parameter space, rather than extending its knowledge boundary.

\begin{figure}[]
    \centering
    \includegraphics[width=0.9\textwidth]{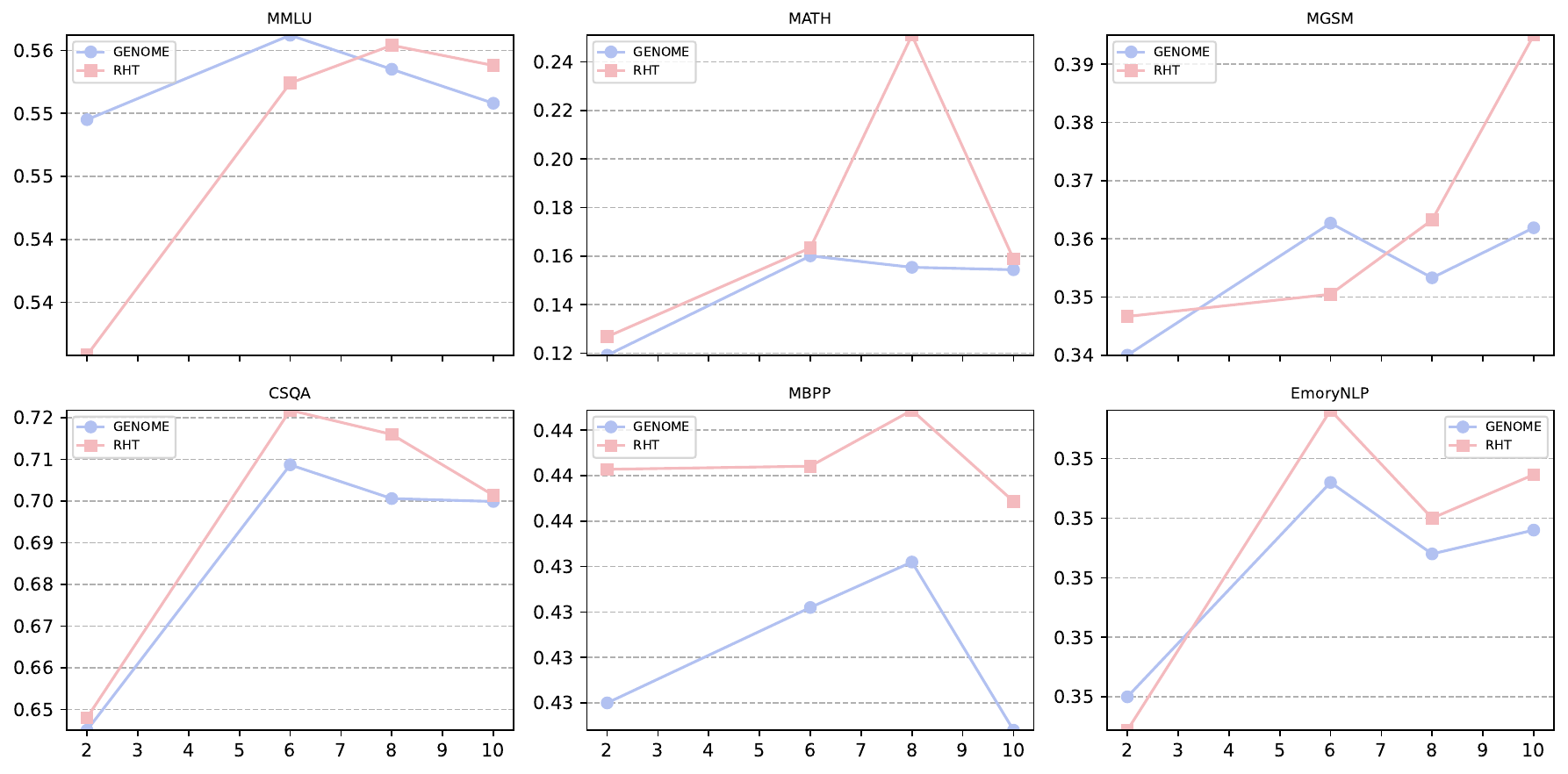}
    \caption{Model merging trends with RHT enhancement.}
    \label{fig:rht}
    \vspace{-1.5em}
\end{figure}

\subsection{Results of Reparameterized Heavy-Tailed Method}\label{sub:3.6}
Figure~\ref{fig:rht} illustrates the performance changes of different merging strategies as the number of merged models increases. The horizontal axis represents the number of merged experts, while the vertical axis represents performance metrics across various tasks, including MMLU, MATH, MGSM, CSQA, MBPP, and EmoryNLP. The results show that RHT significantly improves the number of models merged in several tasks, especially in scenarios where GENOME begins to plateau or degrade. For instance, in tasks like MMLU, MGSM, and MATH, RHT exhibits clear upward trends as more experts are added, outperforming GENOME consistently. This demonstrates RHT’s ability to better utilize the growing number of experts by expanding the effective parameter space through heavy-tailed reparameterization. RHT helps counteract the saturation effect often seen in vanilla model merging. The heavy-tailed design of RHT alleviates this bottleneck by allowing the merged model to explore a wider region of the parameter space, avoiding premature convergence to suboptimal representations. 

\section{Related Works}
\paragraph{Parameter-efficient fine-tuning}Parameter-efficient fine-tuning \cite{li2021prefix,lester2021power,liu2024gpt,liu2024hift} has garnered significant attention in recent years for its ability to adapt pre-trained models to specific tasks by adjusting only a small subset of parameters, thereby significantly reducing computational resource requirements. In the domain of model fusion, PEFT has been widely employed to efficiently integrate multiple expert models, avoiding the complexity of fine-tuning all parameters of each individual model. By optimizing a small number of task-specific parameters, techniques such as expert ensemble \cite{polikar2012ensemble,rame2023rewarded,polyak1992acceleration,zhou2021domain}, low-rank adaptation \cite{hu2022lora,choi2024revisiting,zhang2023composing,luo2024moelora}, effectively enable the fusion of expert models. These methods not only reduce computational costs and memory requirements but also provide a scalable and efficient framework that allows multiple models to be combined into a single, parameter-efficient representation. However, since PEFT primarily focuses on fine-tuning for specific tasks rather than effectively combining multiple expert models, it may fail to fully leverage the strengths of each expert. This creates a trade-off between efficiency and the ability to fully exploit the diverse expertise in model ensembles.

\paragraph{Model Merging}Model merging aims to optimize the merging performance by leveraging the complementary capabilities of different models. Static methods \cite{yadav2023ties,yu2024language} merge model parameters to avoid the need for additional data, while dynamic methods \cite{yang2023adamerging,mavromatis2024pack,prabhakar2024lora} achieve the composition of multiple skills by optimizing the merging weights. To avoid conflicts caused by overlapping subspaces of different tasks, \citet{po2024orthogonal} proposes applying orthogonality constraints during the training phase, while \citet{choi2024revisiting} uses singular value decomposition to separate task-specific knowledge from noise and employs low-rank approximations to reduce task interference. Recent research has modeled the merging of large language models as an optimization problem, with approaches like \cite{akiba2025evolutionary,huang2023lorahub,feng2024model}. However, the former tends to simplify evolutionary mechanisms or focus solely on merging coefficients, while the latter adjusts model weights using swarm intelligence, which may lead to local optima. GENOME \cite{zhang2025nature}, on the other hand, enhances the effectiveness of the evolutionary algorithm by incorporating genetic-level and population-level operations. Despite these efforts to merge multiple expert models, the actual number of experts effectively merged for optimal performance is often much lower than anticipated. To investigate this phenomenon, we start by examining the parameter space of expert models and further expand the parameter space to enable the effective merging of more expert models.

\section{Conclusion}

In this paper, we systematically investigate the fundamental limitations of model merging scalability through rigorous theoretical analysis and empirical evaluation. Our mathematical characterization, grounded in Gaussian Width, reveals an inherent pattern of concave diminishing returns in multi-expert ensembles, attributed to the saturation of the effective parameter space. The derived kinematic threshold provides a theoretical stopping criterion for the merging process. To address these limitations, we propose a reparameterized Heavy-Tailed method that extends the coverage of merging parameters via heavy-tailed geometric reconstruction, resulting in sustained performance improvements.
\section*{Limitations}\label{sec:limitations}
This paper assumes that experts are obtained using LoRA, which may limit the generalizability of its conclusions. For example, the merging behavior of experts fine-tuned with all parameters may not align with the findings presented here.
The theoretical analysis relies on homogeneous model architectures. Real-world scenarios often involve heterogeneous architectures or non-linear interactions between parameters, which may limit the practical applicability of our theories.
\newpage

\bibliography{ref}
\bibliographystyle{plainnat}

\clearpage

\appendix
\section{Experimental Setup}
\label{app:implementation_details}
Our experiments are strictly performed on high-performance computing hardware, NVIDIA-A800-SXM4-80GB, to ensure the efficiency and scalability of the model. To further enhance the reproducibility of the results, we accurately set and record all experimental random seeds, ensuring the exact replication of experimental conditions and outcomes. To obtain expert models, we fine-tune base models on these two types of datasets using LLaMA-Factory \cite{zheng2024llamafactory} with LoRA, following the configurations described in \cite{zhang2025nature}.
For $D_{\text{knowd}}$, we use two base models: LLaMA3.1-8B-Instruct and Gemma-2-2B-it~\cite{team2024gemma}, with model merging performed using GENOME. For $D_{\text{gend}}$, we use Gemma-2-2B-it as the base model and perform model merging using both GENOME and Model Swarms.

\begin{figure}[!h]
  \centering
  \includegraphics[width=\linewidth]{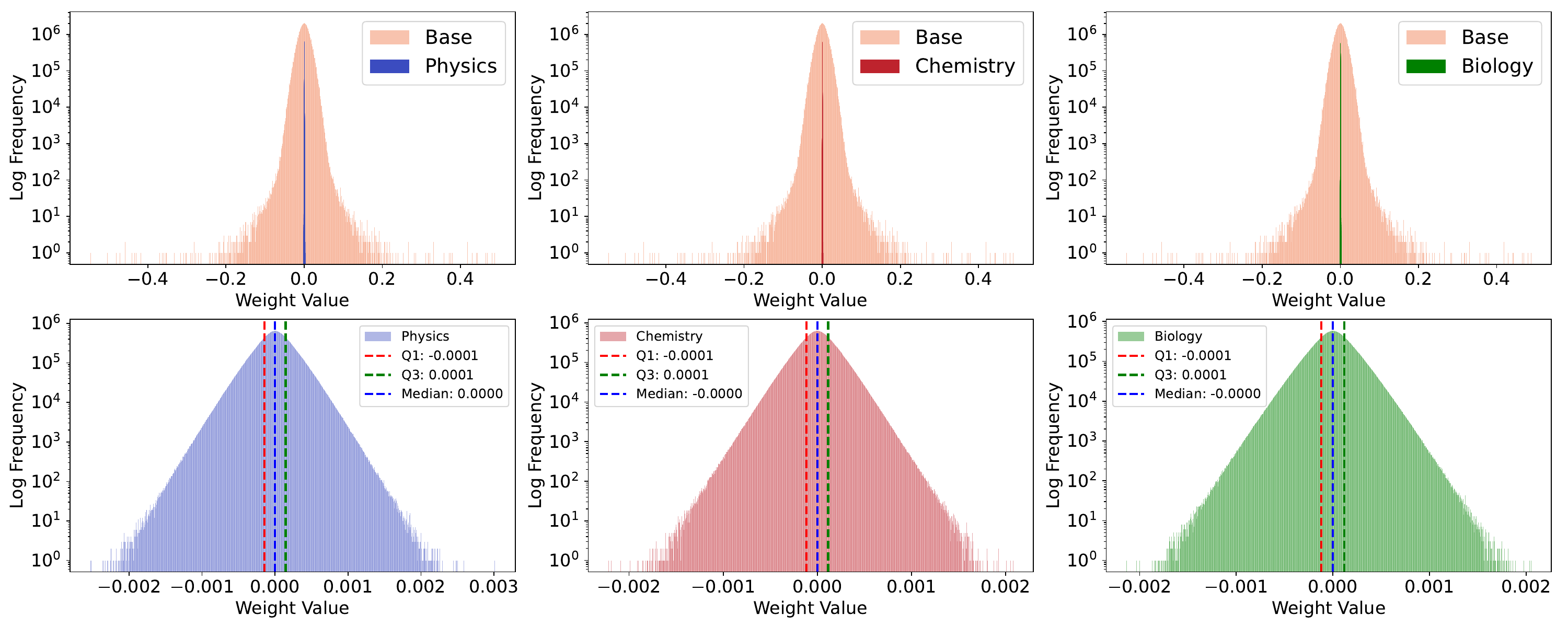}
  \caption{Incremental parameter distribution in LoRA.}
\end{figure}
\begin{figure}[!h]
  \centering
  \includegraphics[width=\linewidth]{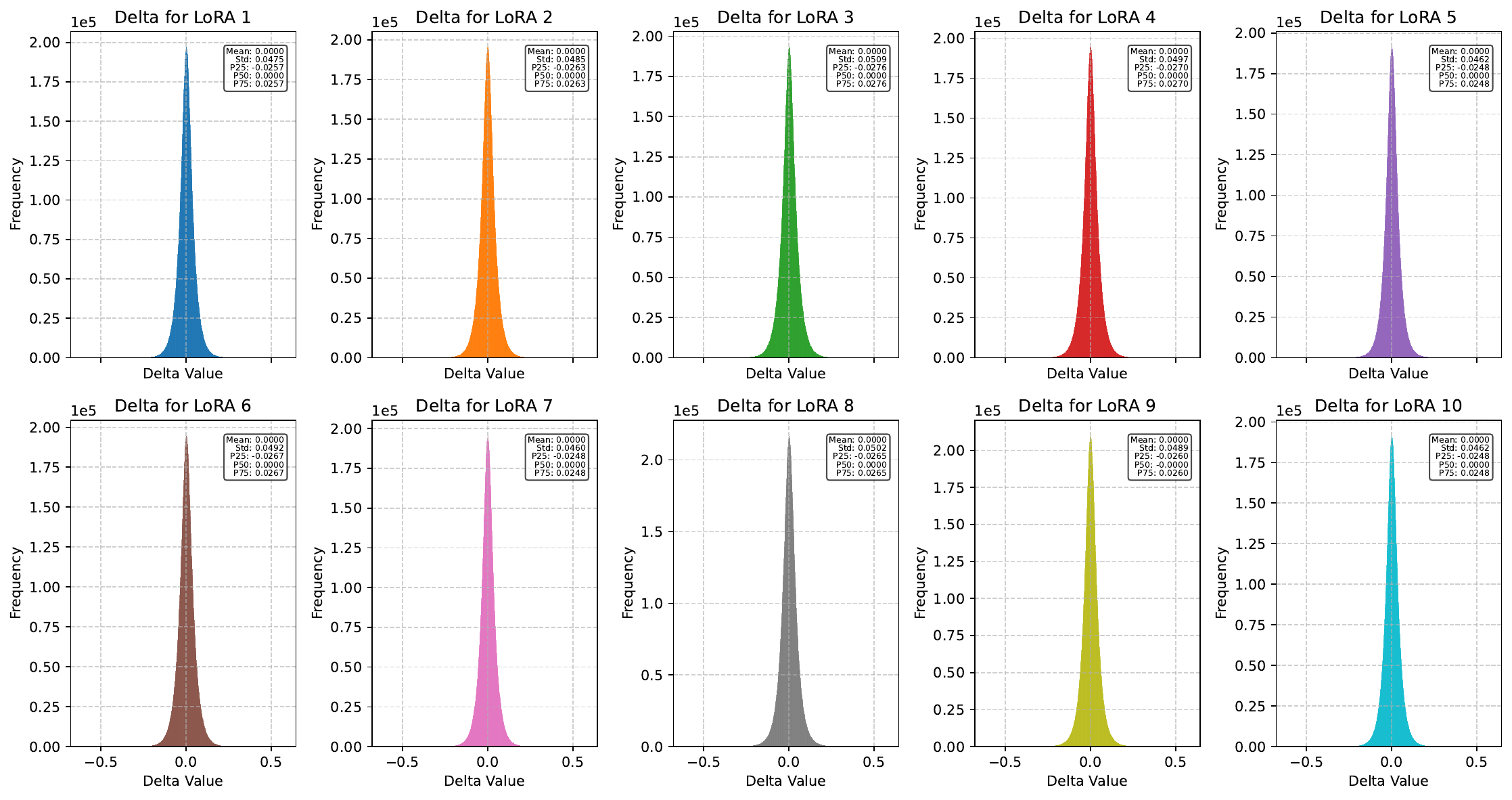}
  \caption{Histogram of expert model weight distributions under model merging.}
  \label{fig:merge weight}
\end{figure}

\section{Datasets}
\label{app:datasets}
We investigate the scalability of model merging across both knowledge-intensive and general-purpose scenarios. The knowledge-intensive setting uses a dataset $D_{\text{knowd}}$, comprising five specialized domains: physics, chemistry, biology, medicine, and finance, while the general-purpose setting uses a dataset $D_{\text{gend}}$, derived from ten diverse domains in the Tulu-v2-SFT-mixture dataset \cite{ivison2023camels}.

\paragraph{General-purpose tasks:}We select six datasets covering six key capabilities of large language models (LLMs), including common sense knowledge (MMLU~\cite{hendrycks2020measuring}), mathematics (MATH~\cite{hendrycks2021measuring}), code generation (MBPP~\cite{austin2021program}),  multilingual processing (MGSM~\cite{shi2022language}), affective computing (EmoryNLP~\cite{zahiri2018emotion}), and question answering (CSQA~\cite{talmor2018commonsenseqa}). Each dataset is split into a $200$-sample validation set and approximately $1,000$ samples for the test set. 
    

\begin{itemize}[leftmargin=*]
\item \textbf{COMMONSENSEQA (CSQA)}~\cite{talmor2018commonsenseqa}: CSQA is a multiple-choice question answering dataset designed to evaluate the AI model’s ability to reason and answer questions based on commonsense knowledge.

\item \textbf{EmoryNLP}~\cite{zahiri2018emotion}: EmoryNLP is a dialogue dataset based on the TV show Friends, containing 97 episodes, 897 scenes, and 12,606 utterances, with each utterance annotated with one of seven emotion categories, i.e., Sad, Mad, Scared, Powerful, Peaceful, Joyful, and a default emotion of Neutral.

\item \textbf{MATH}~\cite{hendrycks2021measuring}: MATH is a dataset that evaluates the mathematical reasoning and problem-solving capabilities of AI models, covering a variety of mathematical problems from basic arithmetic to calculus.

\item \textbf{MBPP}~\cite{austin2021program}: MBPP is a benchmark for evaluating the performance of Python code generation models, covering 974 short Python programming tasks covering topics such as basic programming concepts and standard library usage.

\item \textbf{Multilingual Grade School Math (MGSM)}~\cite{shi2022language}: MGSM is the multilingual version of GSM8K, containing some examples translated into ten languages with different types of languages.

\item \textbf{MMLU}~\cite{hendrycks2020measuring}: MMLU is a benchmark for assessing model performance in zero-shot and few-shot scenarios, testing general knowledge and problem-solving abilities across 57 subjects, and covering multi-task language understanding, question answering, and arithmetic reasoning.
\end{itemize}

\paragraph{Knowledge-intensive tasks:}We design a comprehensive evaluation framework covering physics, chemistry, and biology, which includes two distinct tasks: title generation, and translation. We generate $500$ samples using GPT-4o-mini and manual verification.
Each dataset is split into a $150$-sample validation set and $350$ samples for the test set.

\paragraph{Dataset Construction and Evaluation Scheme:}For $D_{\text{knowd}}$, our dataset construction method begins with systematically randomly selecting 500 seed instances from the original training corpus of the expert models. For each seed instance, we use k-nearest neighbor retrieval from a domain-specific knowledge base to identify semantically aligned reference texts. These retrieved contexts are then processed by GPT-4o-mini to generate task-specific question-answer pairs. Samples that are rejected are iteratively regenerated through consensus scoring by domain experts until they meet the required criteria. The final dataset is split into a validation set (150 instances) and a test set (350 instances). For the biological title generation task, we perform consensus evaluation by domain experts and directly select 200 validation instances and 1077 test instances from the original knowledge base to ensure data provenance. Evaluation strictly follows the zero-shot protocol, without any fine-tuning for specific tasks. For $D_{\text{gend}}$, we use standard benchmark datasets. The detailed splits of the two datasets and the evaluation metrics used are presented in Table \ref{tab:Benchmark}.

\begin{table}[!h]
\caption{Datasets and Evaluation Metrics for Benchmarking.}
\label{tab:Benchmark}
\begin{tabular}{cllm{2.8cm}cc}
\hline
\multicolumn{1}{l}{\multirow{2}{*}{}}& \multicolumn{1}{c}{\multirow{2}{*}{\textbf{Dataset}}} & \multicolumn{1}{c}{\multirow{2}{*}{\textbf{Category}}} & \multicolumn{1}{c}{\multirow{2}{*}{\textbf{Metrics}}} & \multicolumn{2}{c}{\textbf{Size}} \\ \cline{5-6} 
\multicolumn{1}{l}{}& \multicolumn{1}{c}{}& \multicolumn{1}{c}{}& \multicolumn{1}{c}{}& valid& test\\ \hline
\multirow{7}{*}{\begin{tabular}[c]{@{}c@{}}General\\ Purpose\\ Data\end{tabular}}
& CSQA& Question Answering& accuracy, 0-shot& 200& 1000\\
& EmoryNLP& Affective Computing& weighted-F1, 0-shot& 200& 697\\
& MATH& Mathematics& accuracy, 0-shot& 200& 1000\\
& MBPP& Code Generation& Pass@1, 0-shot& 200& 774\\
& MGSM& Multilingual Processing& accuracy, 0-shot& 200& 2637\\
& MMLU& General Knowledge& accuracy, 0-shot& 200& 1000\\ \hline
\multirow{6}{*}{\begin{tabular}[c]{@{}c@{}}Knowledge\\ Intensive\\ Data\end{tabular}} & Physics\_title& Title Generation&BERT Score, F1, ROUGE, BLEU& 150& 350\\
& Physics\_trans& Text Translation&BLEU& 150& 350\\
& Chemistry\_title& Title Generation&BERT Score, F1, ROUGE, BLEU& 150& 350\\
& Chemistry\_trans& Text Translation&BLEU& 150& 350\\
& Biology\_title& Title Generation&BERT Score, F1, ROUGE, BLEU& 200& 1077\\
& Biology\_trans& Text Translation&BLEU& 150& 350\\ \hline
\end{tabular}
\end{table}

\section{Proofs}

\begin{definition}[Gaussian Width~\cite{vershynin2015estimation}]\label{def:Gaussian}
   Let $S \subseteq\mathbb{R}^D$ be a subset of the $D$-dimensional Euclidean space. The Gaussian Width $w(S)$ of $S$ is defined as:
\begin{equation}\label{eq:eq5}
w(S) = \frac{1}{2} \mathbb{E}\left[ \sup_{x, y \in S} \langle g, x - y \rangle \right],
\end{equation}
where $g \sim \mathcal{N}(0, I_D)$ is a standard Gaussian random vector, and $\langle g, x - y \rangle$ represents the inner product between $g$ and the difference $x - y$ between any two points $x$ and $y$ in $S$. 
\end{definition}
The Gaussian Width quantifies the extent to which the set $S$ spans in random directions, thereby reflecting its geometric complexity.
\begin{definition}[Statistical Dimension~\cite{amelunxen2014living}]
    For a closed convex cone $C\subseteq\mathbb{R}^D$, its statistical dimension is expressed as:
    \begin{equation}\delta(C) = \mathbb{E} \left[ \| \Pi_C(g) \|_2^2 \right],\end{equation}
    where $g \sim \mathcal{N}(0, I_D)$ is a standard Gaussian random vector, $\Pi_C(g)$ is the projection of $g$ onto the convex cone $C$.
\end{definition}
\begin{lemma}[Approximate Kinematics Theory~\cite{amelunxen2014living}]\label{lemma:appro}
    For a closed convex cone $C\subseteq\mathbb{R}^D$, any $k$-dimensional subspace $S_k\subseteq \mathbb{R}^D$, and a Haar-distributed random orthogonal matrix $Q$:
\begin{equation}\begin{aligned}
\delta(C)+k\;\lesssim\; D \;\Longrightarrow\; \Pr\{C\cap QS_k=\phi\}\approx1, \\
\delta(C)+k\;\gtrsim\; D \;\Longrightarrow\; \Pr\{C\cap QS_k=\phi\}\approx0.
\end{aligned}\end{equation}\end{lemma}
\subsection{Proof of the Upper Bound Mode Merging}\label{proof:Upper_Bound_o_Merge_Merging}
\begin{proof}[Proof of \Cref{Upper_Bound_o_Merge_Merging}]
According to the linear combination properties of Gaussian random variables, the merge parameter distribution is
\begin{equation}
    \theta_{\mathrm{merge}}
\sim \mathcal{N}(\mu_{\mathrm{merge}},\,\Sigma_{\mathrm{merge}}).
\end{equation}
The mean vector is:
\begin{equation}
    \mu_{\mathrm{fusion}}
= \sum_{i=1}^n \alpha_i \,\mu_i.
\end{equation}
Covariance matrix:
\begin{equation}
    \Sigma_{\mathrm{merge}}
= \sum_{i=1}^n \alpha_i^2\,\sigma_i^2\,I
  \;+\;
  \sum_{i=1}^n \sum_{\substack{j=1\\j\neq i}}^n
    \alpha_i\,\alpha_j\,
    \mathrm{Cov}(\theta_i, \theta_j).
\end{equation}
Define the covariance between experts $i$ and $j$ as
\begin{equation}
    \mathrm{Cov}(\theta_i, \theta_j)
= \rho_{ij}\,\sigma_i\,\sigma_j\,I,
\quad
|\rho_{ij}|\le1.
\end{equation}
Substituting the covariance into the covariance matrix expression above:
\begin{equation}
   \Sigma_{\mathrm{merge}}
= \Bigl(\sum_{i=1}^n \alpha_i^2\,\sigma_i^2
      + \sum_{i=1}^n \sum_{\substack{j=1\\j\neq i}}^n
          \alpha_i\,\alpha_j\,\rho_{ij}\,\sigma_i\,\sigma_j
  \Bigr) 
  I.
\end{equation}
Simplified to scalar variance:
\begin{equation}
    \sigma_{\mathrm{merge}}^2
= \sum_{i=1}^n \alpha_i^2\,\sigma_i^2
  + \sum_{i \neq j} \alpha_i \alpha_j\,\rho_{ij}\,\sigma_i\,\sigma_j.
\end{equation}
The merged variance in the simplified case is:
\begin{equation}
    \sigma_{\mathrm{merge}}^2
= \sigma^2
  \Bigl(\sum_{i=1}^n \alpha_i^2
       + \rho \sum_{\substack{i=1}}^n\sum_{\substack{j=1\\j\neq i}}^n \alpha_i \alpha_j
  \Bigr).
\end{equation}
Noting $(\sum_{i=1}^n \alpha_i)^2 = \sum_{i=1}^n \alpha_i^2 + \sum_{\substack{i=1}}^n\sum_{j=1,i\neq j}\alpha_i\alpha_j = 1$, we get
\begin{equation}
    \sigma_{\mathrm{merge}}^2
= \sigma^2
  \bigl(\rho + (1-\rho)\sum_{i=1}^n \alpha_i^2\bigr).
\end{equation}
In the uniform weight case $\alpha_i = 1/n$, the variance is
\begin{equation}\label{eq:v_sample}
    \sigma_{\mathrm{merge}}^2
=  = \sigma^2\Bigl(\rho + \tfrac{1-\rho}{n}\Bigr).
\end{equation}
When the number of experts $n \to \infty$, the variance after merging tends to:
\begin{equation}
    \lim_{n \to \infty} \sigma_{\mathrm{merge}}^2
= \sigma^2\,\rho.
\end{equation}
This shows that no matter how many models are merged, the variance cannot be lower than $\sigma^2\,\rho$, that is, there is a theoretical lower bound $\sigma^2\,\rho$. When the models are completely independent ($\rho = 0$), theoretically increasing the number of models can reduce the variance infinitely. However, in reality, there is usually a correlation between models ($\rho > 0$), so there is an upper bound to merge.

Because the variance has a lower bound, we hope that the merge variance will be at least one order of magnitude $\Delta>0$ less than the limit value $\sigma^2\rho$.
\begin{equation}
    \sigma_{\mathrm{merge}}^2(n) - \sigma^2\rho
  \;\ge\;\Delta.
\end{equation}
According to Equation~\ref{eq:v_sample}, we can get
\begin{equation}
    \sigma_{\mathrm{merge}}^2(n) - \sigma^2\rho
= \frac{\sigma^2(1-\rho)}{n}
\;\ge\;\Delta
\;\Longrightarrow\;
n \;\le\; \frac{\sigma^2(1-\rho)}{\Delta},
\end{equation}
\begin{equation}
n_{\max}
= \left\lfloor
  \frac{\sigma^2(1-\rho)}{\Delta}
\right\rfloor.
\end{equation}
This indicates that there is an upper bound on the number of models that can be merged, and this upper bound is mainly determined by the correlation between the models.
\end{proof}

\subsection{Proof of the Gaussian Width of the Merged Model Subspace}\label{sub:gaussian width}
\begin{proof}[Proof of \Cref{theorem:1}] The model merging problem can be formulated as the following constrained minimization problem:
\begin{equation}\label{eq:loss}
\min_{M \in \mathbb{Z}^+} M \quad \text{s.t.} \quad \exists  \theta \in S(\epsilon), \quad L(\theta) \leq L(\theta^*) + \epsilon,
\end{equation}
where $ S(\epsilon) $ represents the space of all possible parameter configurations when merging $ M $ experts, defined as:
\begin{equation}\label{eq:eq4}
S(\epsilon) = \{ \theta \in \mathbb{R}^D : L(\theta) \leq L(\theta^*) + \epsilon \}.
\end{equation}
Here, $ \epsilon $ is the performance tolerance threshold. Consider the weights $ \theta^* $ and the loss function $ L(\theta) $ obtained by merging all expert models. In the vicinity of $ \theta^* $, we approximate $ L(\theta) $ using a second-order Taylor expansion. Given that the first derivative of $ L(\theta) $ at $ \theta^* $ is zero, Equation~\ref{eq:eq4} can be reformulated as:
\begin{equation}\label{eq:eq20}
S(\epsilon) = \left\{ \theta \in \mathbb{R}^D : (\theta - \theta^*)^T H (\theta - \theta^*) \leq 2\epsilon \right\},
\end{equation}
where $ H $ is the Hessian matrix of $ L(\theta) $ at $ \theta^* $. Since $ H $ is positive definite, $ S(\epsilon) $ forms an ellipsoid centered at $ \theta^* $.

We then perform a linear transformation $z = H^{\frac{1}{2}} (\theta - \theta^*)$ to express:
\begin{equation}
S(\epsilon) = \left\{ z \in \mathbb{R}^D \mid \|z\|^2 \leq 2\epsilon \right\}.
\end{equation}
From Equation~\ref{eq:eq4}, we have:
\begin{equation}\label{eq:ea9}
\sup_{\theta \in S(\epsilon)} \langle g, \theta - \theta^* \rangle = \sup_{z} \langle g, H^{-\frac{1}{2}} z \rangle \quad \text{s.t.} \quad \|z\|^2 \leq 2\epsilon,
\end{equation}
which is maximized by:
\begin{equation}\label{eq:ea10}
z^* = \sqrt{2\epsilon} \cdot \frac{H^{-\frac{1}{2}} g}{\| H^{-\frac{1}{2}} g \|}.
\end{equation}
Thus, the Gaussian Width becomes:
\begin{equation}
w(S(\epsilon)) = \mathbb{E} \left[ \sqrt{2\epsilon} \cdot \|H^{-\frac{1}{2}} g\| \right].
\end{equation}
By applying Jensen's inequality, we approximate the expected value as:
\begin{equation}
\mathbb{E} \left[ \| H^{-\frac{1}{2}} g \| \right] \approx \sqrt{\text{Tr}(H^{-1})}.
\end{equation}
Hence, the final Gaussian Width is:
\begin{equation}\label{eq:eq13}
w(S(\epsilon)) \approx  \sqrt{2\epsilon \cdot \text{Tr}(H^{-1})}.
\end{equation}
For the number of experts $M$, the Gaussian Width becomes:
\begin{equation}
w(S_M) \approx \sqrt{2\epsilon \cdot \sum_{i=1}^{M} \frac{1}{\lambda_i}}.
\end{equation}
where $\lambda_i$ is the $i$-th eigenvalue of $H$.
The marginal contribution of adding the $ M $-th expert is:
\begin{equation}
\Delta w_M = w(S_M) - w(S_{M-1}) = \sqrt{2\epsilon \cdot \sum_{i=1}^{M} \frac{1}{\lambda_i}} - \sqrt{2\epsilon \cdot \sum_{i=1}^{M-1} \frac{1}{\lambda_i}}.
\end{equation}
Since the square root function is concave, the marginal gain decreases as $ M $ increases:
\begin{equation}
\Delta w_M > \Delta w_{M+1}.
\end{equation}
Thus, diminishing marginal return arises from the concavity of the square root function, leading to progressively smaller contributions from each additional expert to the overall Gaussian Width.
\end{proof}

\subsection{Proof of the Gaussian Width of the Merged Model Subspace}\label{sub:Parameter Redundancy}
\begin{proof}[Proof of \Cref{theorem:2}] Let the weight of the merged model of $ M $ experts be $ \theta^k $, which represents a $ k $-sparse vector containing exactly $ k $ non-zero parameters.
Let $ \theta^* $ be the weight vector obtained by merging all expert models. We decompose $ \theta^* $ into two parts:
\begin{itemize}[leftmargin=*]
    \item $ \theta^k = [\theta_1^*, \theta_2^*, \dots, \theta_k^*] $, which represents the parameters contributed by $ M $ expert models ($ M \leq N $).
    \item $ \theta' = [\theta_{k+1}^*, \theta_{k+2}^*, \dots, \theta_d^*] $, which represents the parameters contributed by the remaining $ N - M $ expert models.
\end{itemize}
Given $ \theta^k $, the sublevel set of the loss function is defined by:
\begin{equation}
S(\theta', \epsilon) = \left\{ \theta' \in \mathbb{R}^{d-k} : L([\theta^k, \theta']) \leq L(\theta^*) + \epsilon \right\}.
\end{equation}
To demonstrate the existence of parameter redundancy in the model merging process, we need to show that there exists a $ \theta^k $ such that the zero vector $ 0 \in \mathbb{R}^{d-k} $ belongs to $ S(\theta', \epsilon) $.

Next, consider the statistical dimension of the projection cone of the set $ S(\theta', \epsilon) $. The statistical dimension of the projection cone is closely related to the geometric structure of the set. Using~\Cref{lemma:appro}, we aim to prove that the statistical dimension of the projection cone of $ S(\theta', \epsilon) $ is full, meaning its dimension is $ d-k $.

Let $ C = p(S(\theta', \epsilon)) $ represent the result of projecting the set $ S(\theta', \epsilon) $ onto the unit sphere $ S^{d-1} $. According to existing research~\cite{amelunxen2014living}, there is the following relationship between statistical dimension and Gaussian Width:
\begin{equation}
w^2(C) \leq \delta(C) \leq w^2(C) + 1.
\end{equation}
Therefore, the relationship between the projected Gaussian Width $ w(p(S(\theta', \epsilon))) $ and statistical dimension is:
\begin{equation}\label{eq:eq32}
w(p(S(\theta', \epsilon)))^2 \gtrsim d - k.
\end{equation}

From Equation~\ref{eq:eq20}, we know that $ S(\theta', \epsilon) $ is an ellipsoid, and all points $ x \in S(\theta', \epsilon) $ are projected onto the unit sphere $ S^{d-1} $, with the projection operation given by:
\begin{equation}
p(S(\theta', \epsilon)) = \left\{ \frac{x - \theta^k}{\| x - \theta^k \|} : x \in S(\theta', \epsilon) \right\}.
\end{equation}
According to Equation~\ref{eq:eq13}, the Gaussian Width of the ellipsoid $ w(S(\epsilon)) $ is approximately:
\begin{equation}
w(S(\epsilon))^2 \approx 2\epsilon \text{Tr}(H^{-1}) = 2\epsilon \sum_{i=1}^d \frac{1}{\lambda_i} = \sum_{i=1} r_i^2,
\end{equation}
From~\cite{larsen2021many}, we modify $ r_i^2 $ to:
\begin{equation}
\frac{r_i^2}{\| \theta^* - \theta_k \|_2^2 + r_i^2}.
\end{equation}
Therefore, the projected Gaussian Width is given by:
\begin{equation}\label{eq:eq24}
w(p(S(\theta', \epsilon)))^2 = \sum_{i=1}^{d-k} \frac{r_i^2}{\|\theta^* - \theta^k\|_2^2 + r_i^2}.
\end{equation}
Here, $ r_i = \sqrt{\frac{2\epsilon}{\lambda_i}} $ is the radius of the ellipsoid, and $ \lambda_i $ is the eigenvalue of the Hessian matrix of the loss function $ L([\theta^k, \theta']) $ with respect to $ \theta' $.

From formulas ~\ref{eq:eq32} and ~\ref{eq:eq24}, it can be observed that as the number of expert models increases, the number of non-zero parameters $ k $ in the network also increases, and the parameter $ \theta^k $ approaches $ \theta^* $, which makes:
\begin{equation}
\frac{r_i^2}{\|\theta^* - \theta^k\|_2^2 + r_i^2} \approx 1.
\end{equation}
In this case, the projected Gaussian Width will approach $ d - k $, that is: $
w(p(S(\theta', \epsilon)))^2 \approx d - k.$
When each fraction $ \frac{r_i^2}{\|\theta^* - \theta^k\|_2^2 + r_i^2} $ approaches 1, it means that the contribution from each direction is close to 1. At this point, the projected Gaussian Width will be close to:
\begin{equation}
w(p(S(\theta', \epsilon)))^2 = \sum_{i=1}^{d-k} 1 = d - k.
\end{equation}
Thus, $ 0 \in S(\theta', \epsilon) $, meaning all the unmerged parameters become redundant.
\end{proof}

\subsection{Proof of the Difference of Gaussian Distributions}\label{sub:Difference of Gaussian Distributions}
\begin{proof}[Proof of \Cref{theorem:4}]
According to the properties of independent Gaussian random variables, their linear combination is still Gaussian, with the mean and variance given by the linear combination of the means and variances, respectively. Therefore,
\begin{equation}
\begin{aligned}
\mathbb{E}[\mathbf{w}']& = \mathbb{E}[\mathbf{w}] - \mathbb{E}[\mathbf{g}] = \boldsymbol{\mu} - \boldsymbol{\mu} = \mathbf{0}, \\
\mathrm{Var}[\mathbf{w}'] = &\mathrm{Var}[\mathbf{w}] + \mathrm{Var}[\mathbf{g}] = \sigma^2 \mathbf{I} + \sigma_g^2 \mathbf{I} = (\sigma^2 + \sigma_g^2) \mathbf{I}.
\end{aligned}
\end{equation}
Thus, $\mathbf{w}' \sim \mathcal{N}(\mathbf{0}, (\sigma^2 + \sigma_g^2) \mathbf{I})$.
\end{proof}
\subsection{Proof of Heavy-Tailed Distribution Induced by Nonlinear Transformation}\label{sub:Nonlinear Transformation}
\begin{proof}[Proof of \Cref{theorem:5}]
For $\mathbf{w}' \sim \mathcal{N}\bigl(\mathbf{0}, (\sigma^2 + \sigma_g^2) \mathbf{I}\bigr)$, the probability density function is
\begin{equation}
p_{\mathbf{w}'}(\mathbf{x}) = \frac{1}{(2\pi (\sigma^2 + \sigma_g^2))^{d/2}} \exp\left(-\frac{\|\mathbf{x}\|^2}{2(\sigma^2 + \sigma_g^2)}\right).
\end{equation}
Define the transformed variable $\mathbf{w}'' = T(\mathbf{w}')$. Using the change of variables formula, for each component $i$, let $y_i = T(x_i)$, and assume $T$ is invertible with inverse $x_i = T^{-1}(y_i)$.

The probability density function of $\mathbf{w}''$ is
\begin{equation}
p_{\mathbf{w}''}(\mathbf{y}) = p_{\mathbf{w}'}\bigl(T^{-1}(\mathbf{y})\bigr) \cdot \left|\det\left(\frac{\partial T^{-1}(\mathbf{y})}{\partial \mathbf{y}}\right)\right|.
\end{equation}
Since $T$ acts component-wise, the Jacobian matrix is diagonal, so
\begin{equation}
\left|\det\left(\frac{\partial T^{-1}(\mathbf{y})}{\partial \mathbf{y}}\right)\right| = \prod_{i=1}^d \left|\frac{d T^{-1}(y_i)}{d y_i}\right|.
\end{equation}
Now, we focus on analyzing the effect of the transformation  
\begin{equation}
T(x_i) = \operatorname{sign}(x_i) \cdot |x_i|^{\gamma} \cdot \bigl(1 + \alpha \cdot e^{-\beta |x_i|}\bigr)
\end{equation}
on the tail behavior of the distribution. When $|x_i|$ is large,  
\begin{equation}
T(x_i) \approx \operatorname{sign}(x_i) \cdot |x_i|^{\gamma},
\end{equation}
because $e^{-\beta |x_i|} \approx 0$.

For $0 < \gamma < 1$, the function $|x|^{\gamma}$ grows rapidly near zero but grows more slowly for large values.

For the inverse function $T^{-1}(y_i)$, when $|y_i|$ is large,  
\begin{equation}
|T^{-1}(y_i)| \approx |y_i|^{1/\gamma}.
\end{equation}
Substituting into the Gaussian density function, when $|y_i|$ is large:  
\begin{equation}
p_{\mathbf{w}''}(y_i) \propto \exp\left(-\frac{|y_i|^{2/\gamma}}{2(\sigma^2 + \sigma_g^2)}\right) \cdot \left|\frac{d T^{-1}(y_i)}{d y_i}\right|.
\end{equation}
Here,  
\begin{equation}
\left|\frac{d T^{-1}(y_i)}{d y_i}\right| \approx \frac{1}{\gamma} |y_i|^{\frac{1}{\gamma} - 1}.
\end{equation}
Therefore,  
\begin{equation}
p_{\mathbf{w}''}(y_i) \propto |y_i|^{\frac{1}{\gamma} - 1} \exp\left(-\frac{|y_i|^{2/\gamma}}{2(\sigma^2 + \sigma_g^2)}\right).
\end{equation}
Since $0 < \gamma < 1$, we have $2/\gamma > 2$, so the power in the exponential term is greater than 2, causing the tail to decay more slowly than a Gaussian distribution.
Moreover, for sufficiently large $|y_i|$, the tail behavior of the cumulative distribution function satisfies:  
\begin{equation}
P(|W_i''| > |y_i|) \sim |y_i|^{-\kappa}.
\end{equation}\end{proof}
\subsection{Proof of Heavy-Tailed Distributions Expanding the Model Function Space}\label{sub:Heavy-Tailed Distributions Expand the Model Function Space}
\begin{proof}[Proof of \Cref{theorem:6}]
Consider two regions in the parameter space $\mathcal{W}$: the central region $\mathcal{W}_C$ and the tail region $\mathcal{W}_T$. For parameters $\mathbf{w} \in \mathcal{W}_T$, i.e., parameters with extreme values, they often induce special nonlinear effects. Specifically, consider a neural network with ReLU activation $\sigma(x) = \max(0, x)$. When some weights take extremely large values, the corresponding neurons exhibit stronger activation or inhibition, producing more diverse functional forms.

Define the mapping $\Phi: \mathcal{W} \to \mathcal{F}$, which maps parameters $\mathbf{w}$ to the corresponding function $f_{\mathbf{w}}$. Then, a volume element $d\mathbf{w}$ in parameter space maps to a volume element in function space given by $\left| \det \left( J_{\Phi}(\mathbf{w}) \right) \right| d\mathbf{w}$, where $J_{\Phi}(\mathbf{w})$ is the Jacobian matrix of $\Phi$ at $\mathbf{w}$.

Under a heavy-tailed distribution, more probability mass in parameter space is concentrated in the tail region $\mathcal{W}_T$. Due to the nonlinear characteristics of neural networks, when parameters lie in $\mathcal{W}_T$, $\left| \det \left( J_{\Phi}(\mathbf{w}) \right) \right|$ is generally large, indicating that a small neighborhood in parameter space maps to a large neighborhood in function space.

The model coverage under the original parameter distribution $p_{\mathbf{w}}$ is
\begin{equation}
\mathcal{C}_1 = \int_{\mathcal{W}} \left| \det \left( J_{\Phi}(\mathbf{w}) \right) \right| p_{\mathbf{w}}(\mathbf{w}) \, d\mathbf{w},
\end{equation}
and the model coverage under the transformed parameter distribution $p_{\mathbf{w}''}$ is
\begin{equation}
\mathcal{C}_2 = \int_{\mathcal{W}} \left| \det \left( J_{\Phi}(\mathbf{w}) \right) \right| p_{\mathbf{w}''}(\mathbf{w}) \, d\mathbf{w}.
\end{equation}
Since $p_{\mathbf{w}''}$ has higher probability density in $\mathcal{W}_T$, where $\left| \det \left( J_{\Phi}(\mathbf{w}) \right) \right|$ is large, the value of the integral $\mathcal{C}_2$ is greater than $\mathcal{C}_1$, i.e.,
\begin{equation}
\mathcal{C}_2 > \mathcal{C}_1.
\end{equation}
This proves that heavy-tailed parameter distributions indeed expand the model coverage.
\end{proof}

\section{More Analysis}
\subsection{Domain Differences in Effective Merge Limits}
Based on the results in Tables \ref{tab:Performance of GENOME and Model} and \ref{tab:one}, the effective merging numbers for $D_{\text{knowd}}$ and $D_{\text{gend}}$ differ. By analyzing the variance of the expert models, we find that the models trained on $D_{\text{gend}}$ exhibit significantly higher variance than those trained on $D_{\text{knowd}}$. For general-purpose tasks, the original models learn a wealth of background knowledge from large-scale datasets, such as web text, encyclopedias, and programming code. This knowledge can be directly transferred across multiple tasks, such as mathematics, programming, and commonsense reasoning. As a result, fine-tuning activates more knowledge, leading to higher variance. In contrast, for knowledge-intensive data, the original models lack sufficient domain-specific knowledge during pretraining, which limits the extent to which fine-tuning can activate the model's capacity, resulting in lower variance.

During the fine-tuning process, changes in variance reflect the degree to which the original model's capacity is activated. Larger variance indicates more significant adjustments to the model parameters, thereby activating more model capabilities. As discussed in~\Cref{theorem:1}, the diminishing marginal effects of Gaussian Width suggest that, as the number of expert models increases, the explainable variance in the parameter subspace also increases, eventually reaching saturation. Therefore, the performance of model merging is not limitless but constrained by the knowledge and variance that the original model possesses.

\newpage
\section{Results}
In this paper, we conduct empirical studies on the $D_{\text{gend}}$ and $D_{\text{knowd}}$ tasks using the following two state-of-the-art methods.
\paragraph{Model Swarms~\cite{feng2024model}}A collaborative search algorithm designed to adapt large language model (LLM) experts using principles of swarm intelligence.   Inspired by Particle Swarm Optimization (PSO), the method treats each LLM as a “particle” navigating the model weight space.   Guided by a utility function and influenced by personal best, global best, and worst checkpoints, these expert models iteratively update their weights and directions to optimize for a target objective.
\paragraph{GENOME~\cite{zhang2025nature}}A population-based evolutionary framework for adapting large language models (LLMs) based on genetic optimization. Inspired by biological evolution, the method treats each LLM as an “individual” with parameters functioning as digital genes. A population of expert models evolves through three key operations: crossover, which merges weights from parent models; mutation, which introduces random perturbations to enhance diversity; and selection, which prioritizes high-performing individuals based on a fitness function.

\begin{table*}[!h]
        \caption{Performance of GENOME and Model Swarms with 2-10 LoRA Fusion on $D_{\text{gend}}$ Corpus. The table shows the results of five runs.}
	\begin{adjustbox}{max width=\textwidth, center}
		\includegraphics[width=\textwidth]{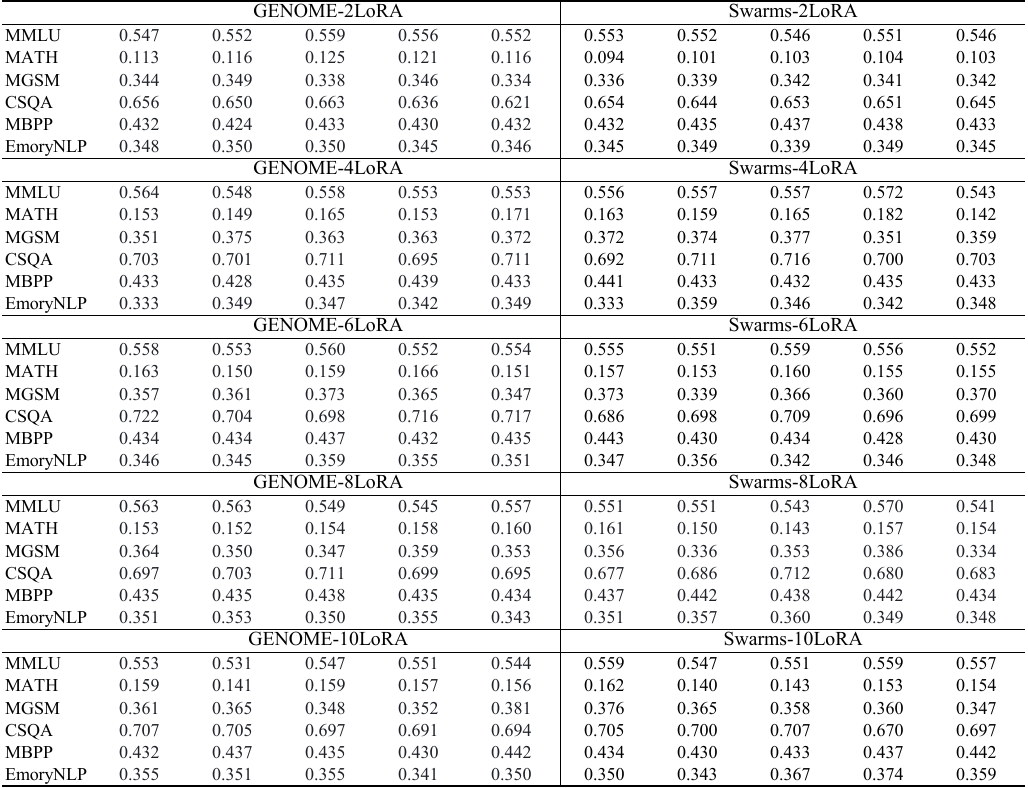}
	\end{adjustbox}
    \vspace{-1em}
	\label{tab:gs_all}
\end{table*}

\begin{table*}[!h]
        \caption{The performance comparison of different LoRA fusion settings on Gemma-2-2B-it across various domain-specific tasks.}
	\begin{adjustbox}{max width=\textwidth, center}
		\includegraphics[width=\textwidth]{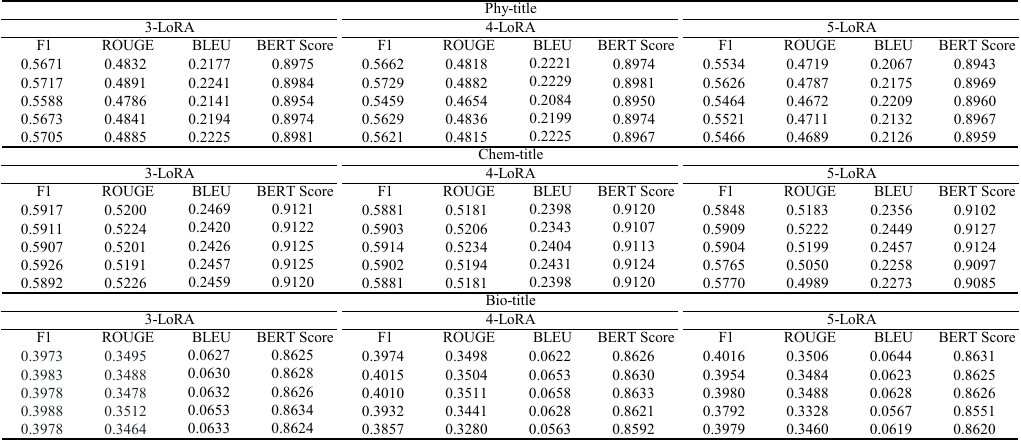}
	\end{adjustbox}
    \vspace{-1em}
	\label{tab:title_gem}
\end{table*}

\begin{table*}[!h]
        \caption{The performance comparison of different LoRA fusion settings on LLaMA3.1-8B-Instruct across various domain-specific tasks.}
	\begin{adjustbox}{max width=\textwidth, center}
		\includegraphics[width=\textwidth]{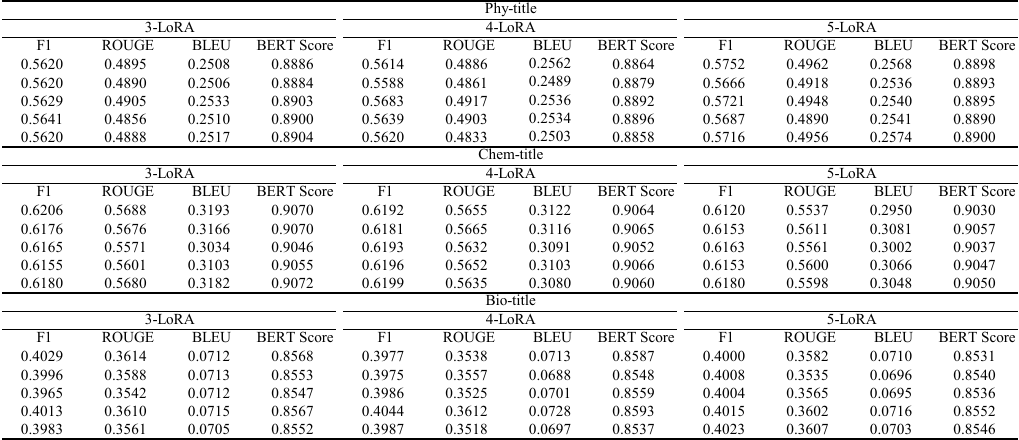}
	\end{adjustbox}
    \vspace{-1em}
	\label{tab:title_llama}
\end{table*}

\begin{table*}[!h]
        \caption{The performance comparison of different LoRA fusion settings on Gemma-2-2B-it and LLaMA3.1-8B-Instruct across various domain-specific tasks. The evaluation metric is BLEU.}
	\begin{adjustbox}{max width=\textwidth, center}
		\includegraphics[width=\textwidth]{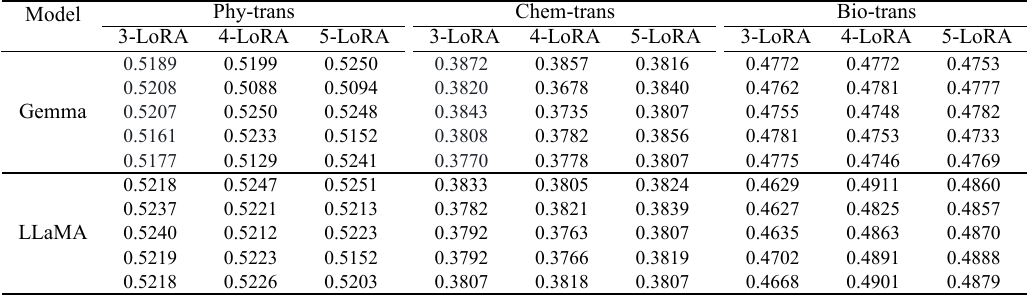}
	\end{adjustbox}
    \vspace{-1em}
	\label{tab:trans}
\end{table*}

\begin{table*}[!h]
        \caption{Performance of pairwise LoRA fusion experiments across domains(Physics).}
	\begin{adjustbox}{max width=\textwidth, center}
		\includegraphics[width=\textwidth]{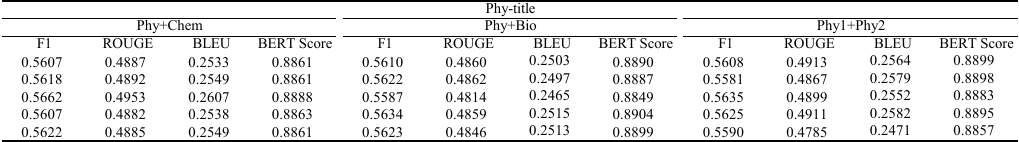}
	\end{adjustbox}
    \vspace{-1em}
	\label{tab:phy_cross}
\end{table*}
\begin{table*}[!h]
        \caption{Performance of pairwise LoRA fusion experiments across domains(Chemistry and Biology).}
	\begin{adjustbox}{max width=\textwidth, center}
		\includegraphics[width=\textwidth]{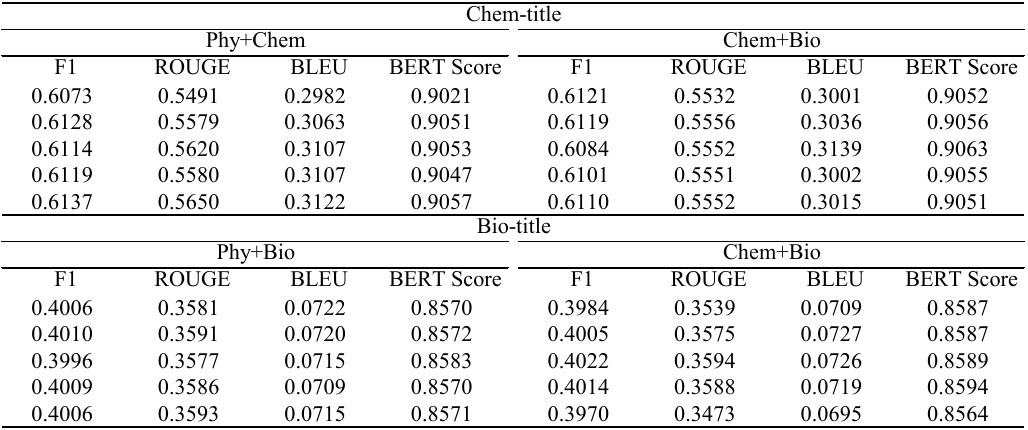}
	\end{adjustbox}
    \vspace{-1em}
	\label{tab:other_title_cross}
\end{table*}
\begin{table*}[t]
        \caption{Performance of pairwise LoRA fusion experiments across domains. The evaluation metric is BLEU.}
	\begin{adjustbox}{max width=\textwidth, center}
		\includegraphics[width=\textwidth]{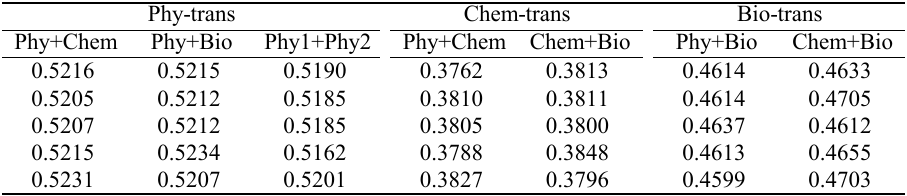}
	\end{adjustbox}
    \vspace{-1em}
	\label{tab:trans_cross}
\end{table*}

\end{document}